\documentclass{article}

\usepackage[final, nonatbib]{neurips_2021}

\usepackage[utf8]{inputenc} % allow utf-8 input
\usepackage[T1]{fontenc}    % use 8-bit T1 fonts
\usepackage{hyperref}       % hyperlinks
\usepackage{url}            % simple URL typesetting
\usepackage{booktabs}       % professional-quality tables
\usepackage{amsfonts}       % blackboard math symbols
\usepackage{nicefrac}       % compact symbols for 1/2, etc.
\usepackage{microtype}      % microtypography
\usepackage{xcolor}         % colors
\usepackage{multirow}
\usepackage{amsmath}
\usepackage{graphicx}
\usepackage{tabto}
\usepackage{enumitem}
\usepackage{subcaption}

\newcommand{\revise}[1]{\textcolor{black}{#1}}

\title{Reliable and Trustworthy Machine Learning for Health Using Dataset Shift Detection}

\author{%
  Chunjong Park, Anas Awadalla, Tadayoshi Kohno, Shwetak Patel\\
  Paul G. Allen School of Computer Science \& Engineering\\
  University of Washington\\
  \texttt{\{cjparkuw, anasa2, yoshi, shwetak\}@cs.washington.edu} \\
}

\begin{document}

\maketitle

\begin{abstract}
%An increasing number of medical diagnostic and screening tools that employ artificial intelligence and machine learning have become available to consumers.
%However, the unpredictable behaviors of ML models on unseen data, especially in the health domain, raise serious concerns about its safety as repercussions for mistakes can be fatal.
Unpredictable ML model behavior on unseen data, especially in the health domain, raises serious concerns about its safety as repercussions for mistakes can be fatal.
In this paper, we explore the feasibility of using state-of-the-art out-of-distribution detectors for reliable and trustworthy diagnostic predictions. 
We select publicly available deep learning models relating to various health conditions (e.g., skin cancer, lung sound, and Parkinson's disease) using various input data types (e.g., image, audio, and motion data). We demonstrate that these models show unreasonable predictions on out-of-distribution datasets.
We show that Mahalanobis distance- and Gram matrices-based out-of-distribution detection methods are able to detect out-of-distribution data with high accuracy for the health models that operate on different modalities.
We then translate the out-of-distribution score into a human interpretable \textsc{confidence score} to investigate its effect on the users' interaction with health ML applications.
Our user study shows that the \textsc{confidence score} helped the participants only trust the results with a high score to make a medical decision and disregard results with a low score. 
Through this work, we demonstrate that dataset shift is a critical piece of information for high-stake ML applications, such as medical diagnosis and healthcare, to provide reliable and trustworthy predictions to the users. 
\end{abstract}

%It might be more familiar for the NeurIPS audience if the abstract focused on a specific contribution rather than an outline of the whole research. For example, I think the NeurIPS audience would most likely to be interested in the ood predictions. 
\vspace{-0.25cm}
\section{Introduction}
\vspace{-0.25cm}
Advances in artificial intelligence and machine learning have made medical diagnostic and screening tools more accurate and accessible.
AI-powered diagnostic tools~\cite{ardila2019end, de2018clinically} are intended to assist medical personnel by making unbiased decision based on thousands of examples.
In recent years, these models~\cite{de2014bilicam, liu2020multi, mariakakis2017biliscreen, huynh2019vitamon} are even becoming available to consumers through the growth of mobile health with the intention of expediting diagnoses through increasingly frequent testing. 
% The tools enable people to do a quick screening of a disease more frequently before its symptoms get more severe.
Moreover, mobile health~\cite{akbar2020safety, Mariakakis2019} aims to improve access to medical expertise for those who are uninsured or live far away from hospitals.

%Despite of its potential benefits, not many consumer facing medical screening tools are approved and available in the market.
Despite the potential benefits of health AI systems, there are concerns about their performance in real-world settings.
Data-driven models learn from examples, making them heavily reliant on the data upon which they have been trained.
However, datasets often fail to get complete coverage over a domain, particularly for emerging datasets; when new pulmonary diseases (e.g., MERS and COVID-19) emerge, a pulmonary classifier trained on the existing lung sounds would not be able to interpret sound of the new diseases.
%for example, a skin leison dataset is heavily skewed toward light colored skin~\cite{} and it will be more difficult for emerging datasets. skin tone color might result in darker AWB, motion blur, etc etc. 
Previous work~\cite{lipton2018detecting, zhang2013domain} has found that machine learning models behave unpredictably on the unseen data.
This problem~\cite{akbar2020safety, subbaswamy2020development} is especially critical for medical diagnostic and screening tools since there are significant repercussions for mistakes.
% The models learn from train dataset that does not sufficiently cover all the input data they would encounter in the real-world settings.
% Unreasonable medical recommendation could be provided to the users when the models encounter an input data drastically different from the train dataset.
% This problem is difficult to address using data-driven approach because it would require infinite number of data.
 
Researchers have proposed methods to estimate the uncertainty of a machine learning models' predictions based on the input~\cite{hendrycks2016a, liang2018enhancing, lee2018, sastry2020detecting, energy2020}.
Out-of-distribution detection methods can distinguish whether the input lies within the distribution of the training dataset, with out-of-distribution data leading to less reliable prediction results.
% is from in-distribution (i.e., similar with dataset used for training) or out-of-distribution (e.g., sufficiently different from dataset used for training), using softmax distribution, Mahalanobis distance, gram, and energy.
% Models' uncertainty is high on out-of-distribution data, resulting in prediction results that are not reliable.
% Thus, the dataset shift provides useful information about the prediction results.
However, such important information has not been widely explored in the context of health applications. 
% used widely for reliability of prediction results, especially in health applications.
When health applications are put into the hands of consumers with limited understanding of the underlying algorithms, they may upload poor quality data that lies outside the distribution of the data that was collected by experts.
For example, consumers who are using a health application that involves image processing may take photographs in poor lighting conditions or framing of the target object.
Even when the data is high-quality, it may be captured with a smartphone that has different hardware specifications than the devices that were used to collect the training dataset.
% e the image One can easily upload input data that the model does not expect (i.e., out-of-distribution data) due to poor instruction (e.g., provide an image of a whole arm instead of a close-up image of a mole on the arm), ambient lighting conditions, and, hardware differences.
Unless the models are explicitly designed or trained to detect invalid data, the models will incorrectly produce a diagnostically meaningless result.
% For example, if an image of a face is fed into the skin cancer classifier that can classify between malignant and benign skin lesions, it will classify the image as either malignant or benign even if a face image is not something that the model can classify for skin cancer.
% Such uncertain results can impact users health decision, which can potentially cause fatal outcome.

In this work, we explore the utility of out-of-distribution detection for improving model performance and user-perceived trustworthiness of health-related models.
We first benchmark our approach using publicly available deep learning models relating to various medical challenges and sensing domains --- images for skin lesion classification, motion data for Parkinson's disease severity, and audio for lung sound classification.
After demonstrating that these models are susceptible to dataset shift, we demonstrate that the state-of-the-art out-of-distribution detectors can effectively exclude such data with over 95\% detection accuracy in most cases.
We then explore the implications of this detector on user-perceived trustworthiness of the health models.
After translating the out-of-distribution score into a human-interpretable metric, \textsc{confidence score}, we found that showing this information to end-users improved the user-perceived trustworthiness of the models.
Furthermore, participants stated that they were more willing to make medical decisions based on models when they were shown the certainty metric.
Our contributions in this work are as follows:
\begin{itemize}
    \item We identify and quantify the limitations of current health deep learning models when encountered with unseen data,
    %\item We design a confidence metric that leverages Mahalanobis distance to quantify the similarity of input data relative to model's training set,
    \item We evaluate the utility of out-of-distribution detection on various data types (e.g., image, audio, motion) for medical screening and diagnosis, and
    \item We evaluate the impact that dataset shift information has on user-perceived trustworthiness of health diagnostic results.
\end{itemize}
\section{Related Work}
%In this section, we describe prior work on (1) ML-based mobile health screening and diagnostic applications, (2) dataset shift detection, (3) trustworthy machine learning. 
%\cj{cut down this section}

\textbf{Machine Learning-based Health Screening and Diagnosis}
In recent years, machine learning has been widely used for medical diagnosis and screening tool to help doctors diagnosis patients easier, faster, and more accurately.
%For examples, there exists machine learning models that can classify lung cancer with x-ray images~\cite{}.
%Machine learning models~\cite{} have been trained to diagnose glaucoma from images of retina.
Machine learning models that learn from large-scale medical datasets are able to detect various symptoms and conditions, including mental health~\cite{gkotsis2017characterisation, su2020deep}, retinal disease~\cite{de2018clinically}, lung cancer~\cite{ardila2019end}.
%A large-scale medical dataset enables the models to outperform human for accuracy and early diagnosis.
With the increasing ubiquity of smartphone and advances in its computing power, machine learning-based health screening can be done on mobile devices.
Various machine learning-based mobile health applications have been proposed to detect health conditions (e.g., traumatic brain injury~\cite{mariakakis2017pupilscreen}, pancreatic cancer~\cite{mariakakis2017biliscreen}, jaundice~\cite{de2014bilicam}) and vital signals (e.g., heart rate~\cite{liu2020multi}, respiratory rate~\cite{liu2020multi}, heart rate variability~\cite{huynh2019vitamon}, blood pressure~\cite{schoettker2020blood}, SpO2~\cite{hoffman2021smartphone}). 
%Pupilscreen~\cite{} segments pupil from eye video and calculates pupil size change over time to screen for traumatic brain injury. 
%VitaMon~\cite{} detects heart rate variability from face video. 
%Seismo~\cite{} uses IMU and camera to estimate people's blood pressure.
%BiliCam~\cite{} and BiliScreen~\cite{} use smartphone camera to detect yellowness from the skin to screen jaundice and pancreatic cancer.
Such mobile health applications can benefit nurses, health workers, and the general population for easier medical screening.
%Nurses and community health workers can also perform medical screening easily with the help of these tools.

While the health machine learning models show high accuracy on their own test datasets, their performance is questionable in real-world settings where the input data can vary drastically, resulting in unreliable prediction results~\cite{thiagarajan2020calibrating, roy2021does}.
Researchers have investigated the dataset shift problem for medical imaging (e.g., x-ray~\cite{cao2020benchmark, ccalli2019frodo}, fundus eye images~\cite{cao2020benchmark}, CT scans~\cite{venkatakrishnan2020self}, dermatology~\cite{roy2021does, pacheco2020out}), focusing on developing and evaluating out-of-distribution detection methods for specific domains. 
However, as more consumer-facing health applications are available in the market, this issue can lead the users to make medical decision based on incorrect results. 
In this work, we aim to explore ways to leverage dataset shift information to make the health machine learning models more reliable and trustworthy to the users.

\textbf{Dataset Shift Detection}
%When machine learning models are deployed in the real-world settings, it is known to fail when encountered with input data that are different from train dataset.
%However, machine learning models fail silently by providing high probability on an incorrect result. The information is limited to inform whether the model provides unreasonable results due to dataset shift. Therefore, machine learning models are notoriously bad at telling the users that they do not know.
%%Unless the models are explicitly trained with ``other'' class, they are designed to classify the input data into one of the predefined classes.
%%Since ``other'' class can be any data that are not relevant to the classes of interest, training with ``other'' class requires a significantly large dataset that covers wide range of ``other'' class data.
%%Consequently machine learning developers often only include classes of interest.
%Even if they include the ``other'' class, the performance of detecting the class is limited unless they can access such large-scale dataset.
Recently researchers have proposed various methods to estimate the models' uncertainty due to dataset shift.
%Unlike the aforementioned data-drive approach to train ``other'' class, 
The proposed methods leverage the output of the models to effectively detect \textit{out-of-distribution} input that are different from the known distribution, \textit{in-distribution}.
Softmax confidence~\cite{hendrycks2016a} has been the baseline for the out-of-distribution detection.
Several work has been proposed for out-of-distribution detection using deep ensemble~\cite{lakshminarayanan2017simple}, Mahalanobis distance~\cite{lee2018}, Gram matrices~\cite{sastry2020detecting}, energy score~\cite{energy2020}, temperature scaling~\cite{liang2018enhancing, sastry2020detecting}, input perturbation~\cite{liang2018enhancing, lee2018}, mean and variance of channels activations~\cite{quintanilha2018detecting}. 
%However, these techniques all require prior knowledge of out-of-distribution dataset which is unlikely scenario in real-world settings.
Alternate training strategies~\cite{hendrycks2018deep, lee2017training, malinin2018predictive, mohseni2020self} have been proposed to enable model to detect out-of-distribution.
Generative models~\cite{nalisnick2019detecting, ren2019likelihood, morningstar2021density, zhang2020hybrid} are proposed to detect out-of-distribution examples.
However, many approaches require re-training and re-designing of the models and prior knowledge of out-of-distribution datasets; it is not realistic to apply these methods to the existing models.
In this work, we explore Mahalanobis distance-~\cite{lee2018}, Gram matrices-~\cite{sastry2020detecting}, and energy-based~\cite{energy2020} out-of-distribution detection methods for reliable and trustworthy machine learning for health since these methods show reasonable out-of-distribution detection performance, do not require retraining or prior knowledge of out-of-distribution datasets, and work on pre-trained discrimitive classifiers.

%In this paper, we employ Mahalanobis distance-based method~\cite{}. Yet, the original work requires prior knowledge of out-of-distribution datasets.
%We modified the existing method to propose the metric that does not require the prior knowledge.

\textbf{Trustworthy AI}
Machine learning systems are deployed in real-world settings to billions of users, making significant impacts on high-stake decision making such as healthcare, policy, economy, and transportation.
Failures in machine learning systems can cause fatal consequences and building trustworthy AI is one of the most important problems in machine learning community.
In recent years, there are active and ongoing efforts aimed at making machine learning systems causal~\cite{athey2017beyond, nie2020quasi}, explainable~\cite{adebayo2018sanity, kim2018interpretability, lundberg2017a, jiang2018to, kaur2020interpreting}, fair~\cite{agarwal2019fair, raghavan2020mitigating, dudik2020assessing}, robust~\cite{evtimov2020security, ford2019adversarial, hendrycks2019benchmarking, eykholt2018robust, tramer2018ensemble}, and privacy-preserving~\cite{abadi2016deep, papernot2018scalable, melis2019exploiting, tramer2021differentially}.
This work contributes to trustworthy AI by improving reliability and user-perceived trustworthiness of machine learning for health using estimated uncertainty.
Bhatt et al.~\cite{bhatt2020uncertainty} proposed to leverage uncertainty for users making decision and placing trust in machine learning models.
This work explores similar approach where we adopt out-of-distribution detection as a method to measure uncertainty.
We took a step further to investigate and quantify its effect in improving reliability and trustworthiness in the context of machine learning for health.
\vspace{-0.20cm}
\section{Background: Dataset Shift Detection Methods}
\vspace{-0.20cm}
% \begin{itemize}
%     \item what is dataset shift
%     \item what dataset shift detection tries to solve
%     \item why we chose mahalanobis distance based ood detection method? briefly answer it with retraining is not necessary, high accuracy, input type agnostic.
%     \item what we will talk about in this section
% \end{itemize}

%Dataset shift occurs when the testing data has differing characteristics to the training data. 
%These differences can be caused by changes in the environment of the model, adversarial inputs, or unseen data. 
%In the context of the health applications, this may manifest itself when a tumor image classifier is exposed to skin tones that were not. part of the training set or when these tumors are present in parts of the body that the classifier has not seen before. 
%Dataset shifts cause the input to be incompatible with the learned features in the model and this leads to a hindrance of model performance and the model can show high confidence in the wrong output. In areas of high stake decision making such as healthcare these models are used to inform diagnosis and treatment decisions. Therefore dataset shifts can have fatal implications and can make machine learning based health models untrustworthy.

%Data shift detection is tasked with distinguishing between in-distribution and out-of-distribution data to allow users to become more informed about the trustworthiness of a model with respect to a specific input. 

%In this work we specifically chose to leverage Mahalanobis distance as this metric ...

We aim to leverage state-of-the-art out-of-distribution detection methods~\cite{lee2018, sastry2020detecting, energy2020} in the health domain for users to safely use health deep learning models.
We selected three out-of-distribution methods that show high accuracy on different out-of-distribution datasets, do not require re-training and prior knowledge of out-of-distribution datasets, and work on pre-trained discrimitive classifiers.
These characteristics are important to help developers or other stakeholders (e.g., regulators, auditors, platforms) easily adopt out-of-distribution detectors to any pre-trained models.
In this section, we provide background on each out-of-distribution detection methods.
%%\cj{pretrained model, no retraining required, good performance, discrimitive classifiers}
%In this section we will present a data shift detection method that uses Mahalanobis distance to detect out-of-distribution samples without the need for prior knowledge of out-of-distribution datasets. Therefore this method can be easily deployed in existing systems with out the need for extensive data collection or retraining of the model as offered by other work in this area.

\subsection{Mahalanobis Distance-Based Out-of-Distribution Detection}
%In Mahalanobis distance-based out-of-distribution detector,
Mahalanobis distance is used to measure the proximity of a point to a certain Gaussian distribution. In the Mahalanobis distance-based out-of-distribution detector~\cite{lee2018}, this property is used to represent each class’s samples at each layer of a network as a class conditional Gaussian distribution with mean $\hat{\mu}_{cl}$ and co-variance $\hat{\Sigma}_{cl}$, where $c$ indicates the class and $l$ indicates the layer in the model. Given a sample input $x$ to a neural network, for each layer, it computes the minimum layer-wise class conditional Mahalanobis distances for $x$. That is, for each layer, it finds the Mahalanobis distance associated with the closest class to $x$. In other words, this is equivalent to $M(x)=\text{max}_c - (f(x) - \mu_c)^{T} \Sigma^{-1} (f(x) - \mu_c)$.
%\begin{align}
%M(x)=\text{max}_c - (f(x) - \mu_c)^{T} \Sigma^{-1} (f(x) - \mu_c)
%\end{align}
The authors have demonstrated that adding small noise to the input can help better distinguish between in-distribution and out-of-distribution data. As the authors suggested, for the real-world setting where the out-of-distribution datasets are generally not available, we obtain the input noise magnitude by generating adversarial samples generated by FGSM~\cite{goodfellow2014explaining}.

%-- Equation format to be fixed --

% In work of Lee et. al~\cite{}, these distances are used as features are used to train a logistic regression model that can distinguish between a specific pair of in-distribution and out-of-distribution datasets. While this method performs well when there is an available out-of-distribution dataset, it is not extendable to real-world scenarios where such dataset is not available. Furthermore, this method does not apply to samples from other out-of-distribution datasets. This prevents it from performing well on the large spectrum of out-of-distribution data present in the wild. 

\subsection{Gram Matrices-Based Out-of-Distribution Detection}
Gram matrices are used to compute pairwise feature correlations and encode stylistic attributes.
For out-of-distribution detection~\cite{sastry2020detecting}, higher order Gram matrices are used to compute class-conditional bounds of feature correlations at all hidden layers of the network as higher order shows better out-of-distribution detection performance. 
Higher order Gram matrices is expressed as $G^P_l = (F_l^PF_l^{P^{T}})^{\frac{1}{P}}$, where $F_l$ is feature map at $l$-th layer and $P$ is order.
%\begin{align}
%    G^P_l = (F_l^PF_l^{P^{T}})^{\frac{1}{P}}
%\label{eq:gram}
%\end{align}
All elements of Gram matrices of an input at each layer are compared against the prepossessed minimum and maximum Gram matrices element values from in-distribution dataset to obtain deviation.
If the input data is predicted as a certain class, the minimum and maximum values of the corresponding class will be used for comparison.
The comparison is done for each layer to obtain layerwise deviations.
Then, the deviations are used to get a total deviation, which is defined by the normalized sum of layerwise deviations.
Whether the input data is from out-of-distribution is determined with a threshold which is defined as 95\% percentile of the total deviations of in-distribution energy score distribution.

\subsection{Energy-Based Out-of-Distribution Detection}

The energy-based out-of-distribution detector~\cite{energy2020} seeks to provide an alternative scoring function to the softmax function that is less susceptible to over-confidence and therefore can better distinguish between in and out-of-distribution inputs. It takes a discriminative classifier $f(c)$ that maps input $x \in \mathbb{R}^D$ to logits, which are traditionally used to derive a categorical confidence score using a softmax function. It defines the energy function on the classifier as $E(x;f) = -T \cdot log \sum_i^K e^{f_i(x)/T}$, where $K$ is the number of classes in the model's output space and $T$ is a temperature parameter that can be used to alter the shape of the energy score distribution. 
%\begin{align}
%    E(x;f) = -T \cdot log \sum_i^K e^{f_i(x)/T}
%\label{eq:energy}
%\end{align}
%In-distribution samples often has the energy score that is more negative compared to out-of-distribution scores. 
Energy score threshold that distinguishes between in- and out-of-distribution samples is defined at the 95\% percentile of the in-distribution samples.

\vspace{-0.25cm}
\section{Performance Evaluation}
\vspace{-0.25cm}
\label{sec:tech-study}
In this section, we demonstrate the performance degradation of the existing health models when encountered with out-of-distribution datasets highlighting that the existing models are vulnerable to dataset shift. 
%The goal is to investigate the feasibility of using out-of-distribution detection to provide more reliable predictions. 
We then evaluate the performance of state-of-the-art out-of-distribution detectors for distinguishing between in- and out-of-distribution examples.
We have selected out-of-distribution datasets that consist of both near- and far-from-distribution samples that represent realistic use cases in the real-world settings.
\revise{For mobile health applications that use mobile sensors for health screening, non-expert users are expected to input data collected by themselves.
%We believe this is where models are most susceptible to out-of-distribution inputs, providing unreliable predictions to the users. 
Unlike clinicians, who may either receive training on how to operate these mobile apps or may already understand what must be done to generate high-quality, non-expert consumers may collect relevant but low-quality data due to environmental factors or totally irrelevant data by mistake or a lack of understanding. 
To reflect these scenarios, we include out-of-distribution datasets caused by covariate shift, label shift, and open-set recognition. 
The covariate and label shifts aim to evaluate a model's performance when tested on data pertaining to the same classification task but from different data sources and environment. 
Open-set recognition evaluates the model's performance on new classes not included in the training set. 
In Table~\ref{tab:ood-performance}, we indicate dataset shift type for each out-of-distribution dataset.}

\subsection{Models and Datasets}
\label{sec:models-datasets}
\textbf{Skin lesion }
A DenseNet-121 based skin lesion classifier~\cite{pacheco2020out} was used in this work.
The model aims to classify an image into seven different skin lesions: actinic keratoses, basal cell carcinoma, benign keratosis, dermatofibroma, melanoma, melanocytic nevi and vascular lesions.
The following datasets are used for training and evaluation: 
\begin{itemize}
    \item (\textbf{In-distribution}) HAM10000~\cite{tschandl2018ham10000, codella2019skin}: (CC BY-NC 4.0) A dataset containing 10,000 samples of dermatoscopic skin tumor images taken using different devices and from different populations. These tumors are part of 7 classes: actinic keratoses, basal cell carcinoma, benign keratosis-like lesions, dermatofibroma, melanoma, melanocytic nevi, and vascular lesions. 
    \item (\textbf{Out-of-distribution}) ISIC 2017~\cite{codella2018skin}: (CC BY-NC 4.0) A previous version of the HAM100000 dataset which contains ~2000 dermatoscopic skin tumor images labelled for binary classification. A tumor is labelled malignant if it corresponds to melanoma to benign if it corresponds to nevus or seborrheic keratosis.
    %\item (\textbf{Out-of-distribution}) ISIC 2016~\cite{gutman2016skin}: (CC BY-NC 4.0) A previous iteration of the HAM100000 dataset which contains ~1300 dermatoscopic skin tumor images labelled malignant or benign.
    \item (\textbf{Out-of-distribution}) Face~\cite{DeBruine2021}: (CC BY 4.0) A dataset containing frontal view face images of 102 adults without making a neutral facial expression. Face images are personally identifiable information. But, all individuals gave signed consent for their images to be ``used in lab-based and web-based studies in their original or altered forms and to illustrate research (e.g., in scientific journals, news media or presentations).''
    \item (\textbf{Out-of-distribution}) CIFAR-10~\cite{krizhevsky2009learning}: (MIT License) A common image classification benchmark with 10 non-medical classes (airplane, car, cat, dog, horse, bird, deer, ship, frog, truck) which contains 6,000 images per class.
\end{itemize}

\textbf {Lung Sound}
A lung sound classification model~\cite{gairola2020respirenet} classifies normal lung sound, wheeze, and crackle from an audio sample. This model is based on ResNet-34 and uses spectograms of audio samples as inputs and outputs 4 lung sound classes (normal, wheezing, crackle, and wheezing + crackle).

\begin{itemize}
    \item (\textbf{In-distribution}) ICBHI 2017 Respiratory Challenge~\cite{rocha2019open}:
    A dataset collected using multiple microphones and stethoscopes containing 6898 samples normal lung sound, wheeze, and crackle audio
    \item (\textbf{Out-of-distribution}) Stethoscope~\cite{stetho2021}: (CC BY 4.0) A dataset containing stethoscope respitory sounds with 336 samples of normal breathing, wheeze, and crackle audio sounds. The dataset was collected using a 3M Littmann Electronic Stethoscope.
    \item (\textbf{Out-of-distribution}) AudioSet~\cite{audioset2017}: (CC BY 4.0)
    A large dataset of millions of sound labelled YouTube audio of which a portion of the dataset contains breathing, cough, and wheezing samples which we use to create a suitable out-of-distribution dataset for this model. 
\end{itemize}

\textbf{Parkinson's Disease}
This is a binary classification model~\cite{zhang2020deep} that showed highest performance in Parkinson's disease digital biomarker DREAM challenge~\cite{sieberts2021crowdsourcing}. The model uses accelerometer signals to detect tremors in a person's movement and outputs whether a participant has Parkinson's. This model consists of 5 1D-convolutional layers and a single output.
\begin{itemize}
    \item (\textbf{In-distribution}) mPower~\cite{bot2016mpower}: (CC BY 4.0) A dataset contains 30-second accelerometer readings from 3,100 participants at rest for both healthy and Parkinson's patients. The dataset was used in Parkinson's disease digital biomarker DREAM challenge~\cite{sieberts2021crowdsourcing}.
    \item (\textbf{Out-of-distribution}) Kaggle Parkinson's disease~\cite{giorgia_2018}: (CC0 1.0) A dataset with accelerometer readings from healthy participants simulate movements of Parkinson's patients.
    \item (\textbf{Out-of-distribution}) MotionSense~\cite{motionsense2019}: (MIT License) A dataset contains accelerometer readings from 24 participants performing various activities (e.g., walking, jogging, sitting, standing, etc).
    \item (\textbf{Out-of-distribution}) MHEALTH~\cite{banos2014mhealthdroid}: An activity classification dataset which contains accelerometer readings from 10 participants executing various activities (e.g., standing, sitting, walking, cycling, etc).
    %as well as the corresponding activity that subject was practicing.
\end{itemize}

\subsection{Performance Impact by Dataset Shift}
\label{sec:performance-eval-setup}

%\subsubsection{Experimental Setup}
In evaluating the model's performance on the out-of-distribution dataset, we used pre-trained models from the previous work\footnote{\url{https://github.com/microsoft/RespireNet}}~\cite{gairola2020respirenet} when the authors make it available. Otherwise, we trained the model in the same way specified in their work\footnote{\url{https://github.com/GuanLab/PDDB}}~\cite{pacheco2020out, zhang2020deep}. We trained skin lesion model~\cite{pacheco2020out}  for 150 epochs using Adam optimizer with a learning rate of 0.0001 and weight decay of 0.2. %batch size of 40
For Parkinson's disease model~\cite{zhang2020deep}, we trained for 50 epochs using Adam optimizer with a learning rate of 0.0005. The pre-trained lung sound model~\cite{gairola2020respirenet} is trained for 200 epochs using SGD optimizer with a learning rate of 0.001 and momentum of 0.9. %with batch size of 64 
For all of these models, we used an 80/20 split and applied the same preprocessing for train and test sets. 
All training and testing is done in a server (Intel Xeon 2.1GHz, 64GB, GeForce RTX 2080 Ti) from an internal cluster.
%We also applied the same normalization and preprocessing used in training. This is done to ensure that the experiment simulates the way the machine learning applications are used in the real-world settings and the variability between in- and out-of-distribution datasets is due to the input features. 
We then ran inference on each dataset and calculated the classification accuracy for the datasets that have corresponding labels. For the datasets that do not have the same labels from the in-distribution, the accuracy could not be computed. Table~\ref{tab:model-performance} summarizes the classification accuracy for the models on in- and out-of-distribution datasets.
%For the skin cancer model there is no meaningful target prediction for the Face and CIFAR10 dataset and therefore we don't include an accuracy score on these datasets. Similarly for the Parkinson's disease model we assume that the target prediction for the activity datasets, MotionSense and mHealth, is negative Parkinson's.

% \usepackage{multirow}
\begin{table}[!t]
\small
\centering
\begin{tabular}{llllll}\toprule
\textbf{Health ML Models} & \textbf{In-Distribution} & \multicolumn{3}{c}{\textbf{Out-of-Distribution}} \\\midrule
%Skin Lesion & HAM10000 & ISIC 2017 & ISIC 2016 & Face & CIFAR \\
Skin Lesion & HAM10000 & ISIC 2017 & Face & CIFAR\\
(DenseNet-121) & 92.05\% & 74.00\% & N/A & N/A \\\midrule
Lung Sound & ICBHI 2017 & Stethoscope & AudioSet  &  \\
%(ResNet-34) & 78.50\% & 2.10\% & 17.31\% &  \\\midrule
(ResNet-34) & 78.50\% & 2.10\% & N/A &  \\\midrule
Parkinson's Disease & mPower & Kaggle Parkinson's & MotionSense & MHEALTH \\
(5$\times$1D-Conv) & 82.01\% & 26.67\% & 45.83\% & 10.00\%  \\\bottomrule
\end{tabular}
\caption{Accuracy of health deep learning models on in-distribution and out-of-distribution dataset. Accuracy is not available (N/A) for out-of-distribution datasets that do not have corresponding labels.}
\label{tab:model-performance}
\end{table}

We generally observed a significant performance drop for all health machine learning models that are tested with out-of-distribution datasets. 
The models output unreasonable and arbitrary predictions on datasets that are not related health conditions.
For example, skin lesion classifier predicts all face images as vascular lesions and CIFAR10 images as various types skin lesions.
Similarly, Parkinson's disease classifier predicts significant portion of physical activities by health participants as tremor caused Parkinson's disease.
For lung sound classification, ordinary sound events (e.g., speech, walking, laughing) are classified as a certain type of lung sounds (e.g., crackle, wheezing).
When the models are evaluated on out-of-distribution datasets that have similar data characteristics to the in-distribution data(i.e., near-distribution datasets), all health models exhibit a performance decrease that ranges from 18\% to 76\%.
This implies that the models are also sensitive to small dataset shift, such as datasets collected with different devices and in different environments.
All of these failure scenarios can occur in real-world deployment of health machine learning applications.
Users can input a face image to skin lesion classifier, improperly record lung sound and input ambient sound to the lung sound classifier, or input motion data when they are not at rest to the Parkinson's disease classifier.
Furthermore, diverse sensors and devices used in real-world deployment can cause significant performance drop.
This evaluation demonstrates that users are exposed to the health machine learning applications that can provide unreliable diagnostic results.

\subsection{Out-of-Distribution Detection Performance}
\label{sec:ood-study}
%Although all models have shown acceptable performance on in-distribution data, the models generally showed performance degradation for both near- and out-of-distribution. 
%This evaluation implies that the users exposed to unreliable medical diagnostic results when the models are deployed in real-world settings. 
%This also shed a light on the importance of distinguishing input data that the models are uncertain.
The previous evaluation implies that it is crucial to determine whether the input data belongs to in- or out-of-distribution to avoid failures caused by dataset shift.
In this section, we investigate the feasibility of using state-of-the-art out-of-distribution detection methods in the context of machine learning for health.
We evaluate Mahalanobis distance-~\cite{lee2018}, Gram matrices-~\cite{sastry2020detecting}, and energy-based~\cite{energy2020} methods, \revise{which work on any pre-trained discrimitive classifiers and do not need re-training and prior knowledge of out-of-distribution datasets}, in detecting out-of-distribution data for different health models.

\subsubsection{Experimental Setup}
\label{sec:ood-setup}
For Mahalanobis distance-based method\footnote{\url{https://github.com/pokaxpoka/deep_Mahalanobis_detector}}, we extracted Mahalanobis distance-based scores from the output dense and residual block found in DenseNet and ResNet respectively. For the Parkinson's model which does not contain dense and residual blocks, we extracted the scores at the end of each convolutional layer.
Then, we optimized the input noise magnitude using in-distribution samples and corresponding adversarial samples generated by FGSM~\cite{goodfellow2014explaining}.
The noise magnitude obtained is 0.0 for skin lesion classifier, 0.0005 for lung sound classifier, and 0.0 for Parkinson's disease classifier.
For Gram matrices-based method\footnote{\url{https://github.com/VectorInstitute/gram-ood-detection}}, we extracted class-specific minimum and maximum correlation values for all orders of Gram matrices for all feature pairs. Total deviation values, which are used for out-of-distribution detection threshold, are computed with multiple sets of random samples from in-distribution datasets.
For energy-based method\footnote{\url{https://github.com/wetliu/energy_ood}}, we use their method that does not require fine-tuning to avoid re-training of the network. 
We use the default temperature scaling ($T=1$) as suggested in ~\cite{energy2020}.
All evaluations are repeated for 5 trials and we report the mean (Table~\ref{tab:ood-performance}) and standard deviation (Table~\ref{tab:ood-performance-ci}) of all metrics.

%setup our experiments for out-of-distribution detection by applying the same normalization and preprocessing used in training to the out-of-distribution datasets. This allows us to make sure that the variability between in and out-of-distribution datasets is due to the features of the input. 

%The mahalanobis distance and gram metric out-of-distribution detectors require the extraction of features from intermediate forward passes in the model. Particularly this can be done by extracting the features in a layer wise fashion or in the case of ResNet and DenseNet architectures extracting features after each residual block or dense block respectively. We find that for the skin cancer (DenseNet-121) and lung sound (ResNet-34) models extracting the features in a block wise manner yields the best results. On the other hand, the Parkinson's disease model doesn't have block components and therefore we extract features from that model layer wise. 

\subsubsection{Evaluation Metrics}
For out-of-distribution detection, it is common to use true negative rate (TNR) at 95\% true positive rate (TPR), AUROC, and detection accuracy to evaluate the performance of a detector. Particularly, as the out-of-distribution problem is a binary classification problem, we consider out-of-distribution samples as negative and in-distribution samples as positive. TNR at TPR 95\% is defined as the percentage of correctly detected out-of-distribution samples, when 95\% of in-distribution samples are correctly detected. The AUROC metric measures the area under the TPR vs FPR curve. The detection accuracy measures the maximum possible classification accuracy over all possible thresholds in distinguishing between in-distribution and out-of-distribution examples. Detailed explanations on the metrics are available in Appendix~\ref{sec:perf-metrics}.

\subsubsection{Results}

\begin{table}[!t]\centering
\resizebox{\textwidth}{!}{
\begin{tabular}{lrrrrrrr}\toprule
\textbf{} &\textbf{} &\textbf{} &\textbf{} &\multicolumn{3}{c}{\textbf{Validation on OOD Samples (TNR @ TPR95/AUROC/Detection Accuracy)}} \\\cmidrule{5-7}
\textbf{Health ML Models} &\textbf{In-Distribution} &\textbf{Out-of-Distribution} &\textbf{Distribution Shift} &\textbf{Mahalanobis} &\textbf{Gram} &\textbf{Energy} \\\midrule
%Skin Cancer &HAM10000  &HAM10000 Test Set &4.57 / 48.69 / 50.43 &0.000 / 18.063 / 55.183 &8.069 / 57.565 / 56.369 \\
Skin Lesion &HAM10000  &ISIC 2017 & Covariate/label shift &10.13 / 58.21 / 59.28 &25.90 / 81.14 / 74.98 & 14.28 / 76.20 / 70.76 \\
%(DenseNet-121) & &ISIC 2016 &4.67 / 51.49 / 53.66 &10.06 / 58.93 / 58.30 & 7.80 / 54.60 / 53.97 \\
(DenseNet-121) & &Face & Open-set recognition &100.00 / 99.98 / 99.96 &95.01 / 98.20 / 96.34 & 0.00 / 80.45 / 84.81 \\
& &CIFAR10 & Open-set recognition &99.83 / 99.90 / 99.61 &95.14 / 98.66 / 96.90 & 5.06 / 58.33 / 57.89 \\\midrule
%Lung Sound &ICBHI &ICBHI Test Set &4.88 / 49.02 / 50.43 &5.035 / 49.578 / 50.236 &5.194 / 53.454 / 54.238 \\
Lung Sound &ICBHI &AudioSet & Open-set recognition & 97.96 / 99.47 / 97.34 & 96.55 / 99.18 / 95.97 &8.12 / 56.79 / 57.13 \\
(ResNet-34) & & Stethoscope & Covariate/label shift & 45.60 / 86.27 / 80.57 & 41.77 / 83.75 / 76.05 &  7.29 / 60.98 / 58.94  \\\midrule
%Parkinson's Disease &mPower &mPower Test Set &3.75 / 50.47 / 51.33 &19.418 / 36.309 / 61.995 &4.800 / 50.794 / 51.016 \\
Parkinson's Disease &mPower &MotionSense & Open-set recognition &100.00 / 99.86 / 99.89 &100.00 / 99.94 / 99.60 & 0.00 / 58.71 / 64.96 \\
(5$\times$1D-Conv) & &mHealth & Open-set recognition &100.00 / 100.00 / 100.00 &100.0 / 99.99 / 99.99 & 0.00 / 41.41 / 59.44 \\
& &Kaggle Parkinson's & Covariate/label shift & 98.00 / 99.89 / 99.47 & 98.96 / 99.96 / 99.67 & 70.00 / 95.91 / 93.34 \\\bottomrule
\end{tabular}}
\caption{Out-of-distribution detection performance across different networks and datasets. }
\label{tab:ood-performance}
\end{table}
Table~\ref{tab:ood-performance} shows out-of-distribution detection performance for different methods across different health machine learning models and datasets. 
Overall, Mahalanobis distance- and Gram matrices-based out-of-distribution detection methods consistently show outstanding performance across different neural networks and different out-of-distribution datasets, showing TNR @ TPR95 of 95\% or above.
These methods show lower performance in distinguishing near-distribution datasets (e.g., ISIC 2017, Stethoscope, Kaggle Parkinson's), which aligns with the results from previous out-of-distribution work~\cite{sastry2020detecting}.
On the other hand, the energy-based method did not show reasonable performance in detecting out-of-distribution samples.
\revise{We found that the energy scores of out-of-distribution samples were not able to effectively discriminated from in-distribution samples as shown in Appendix~\ref{sec:energy-analysis}. Note that we used the energy scores without fine-tuning; however, the authors of energy score-based method~\cite{energy2020} have demonstrated that a classifier that is fine-tuned using the energy score in place of the softmax score shows significant improvement in out-of-distribution detection performance.}
This evaluation implies that state-of-the-art out-of-distribution detectors can be applied to health machine learning applications to provide reliable diagnostic results to the users.

\vspace{-0.25cm}
\section{User Study}
\vspace{-0.15cm}
%The evaluation of out-of-distribution detection demonstrates that it can effectively distinguish between in- and out-of-distribution data.
%From the model's perspective, the system can choose not to process the input data in the background and wait for new input to provide more accurate prediction.
%However, it is still unknown how users would react to the health predictions by the models.
% We show a feasibility of using out-of-distribution detection to filter out unreliable prediction results, dataset shift information can enhance human-AI interaction for health machine learning applications.
According to the trustworthy AI literature~\cite{felzmann2019transparency}, providing users with interpretable information can enhance the trustworthiness of the result and potentially impact users' decisions.
We therefore conducted an online survey-based user study to validate this effect and the impact that our approach has on model trustworthiness.
We first defined \textsc{confidence score} as how confident the model is in interpreting an input.
In other words, in-distribution input would have a high \textsc{confidence score}, whereas out-of-distribution input would have a low \textsc{confidence score}.
\revise{We compute \textsc{confidence score} by scaling raw out-of-distribution scores from out-of-distribution detectors~\cite{lee2018, sastry2020detecting} to 0--100, where 0 is most likely to be an out-of-distribution example and 100 is most likely to be an in-distribution example. 
Scaling is done in a piecewise manner. When out-of-distribution scores are within an in-distribution threshold, which is set to include 95\% of in-distribution examples, we compute min-max scaling that ranges from 90 to 100. In this way, we ensure that most of the in-distribution examples have confidence scores of 90 or above. When out-of-distribution scores outside of an in-distribution threshold, we compute min-max scaling from 0 to 90, where the same denominator is used as above since out-of-distribution examples might not be available in practice and any negative values are clipped to 0.}
We then investigate the effect of \textsc{confidence score} on user-perceived trustworthiness and its impact on medical decisions. Additionally, we also quantify potential learning that can be gained when it comes to distinguishing between in- and out-of-distribution input samples.
Specifically, we aim to answer the following research questions:
\begin{enumerate}[label=RQ.~\arabic*]
    \item How does the \textsc{confidence score} affect the perceived trustworthiness of diagnostic results?
    \item How does the \textsc{confidence score} affect medical decisions based on diagnostic results?
    \item Is there a potential learning effect from \textsc{confidence score} when it comes to distinguishing between input data with high and low \textsc{confidence score}?
\end{enumerate}

\subsection{Study Procedure and Participants}
\label{sec:user-study-procedures}
The overview of the online user study is illustrated in Figure~\ref{fig:user-study} and a list of the user study interfaces is detailed in Appendix~\ref{sec:user-study-interface}.
In short, the interface displays simulated results from the health screening models used in Section~\ref{sec:tech-study} (i.e., models for skin cancer, lung sound, and Parkinson's disease).
We made the input data as human-readable as possible to maximize interpretability.
Images were shown as is, while audio was included in an audio player so that participants could play, pause, and stop the track.
% Because the models receive different input data type (e.g., image, audio, and motion data), we ensure that the participants can correctly interpret and understand what it means.
% We clearly display the input image and embed an audio player for playback.
% Motion data can be difficult to understand by the participants. 
We presented the motion data as a time-series plot of accelerometer signals from x-, y-, and z-axis.
Since time-series data can be particularly challenging for non-experts, we explain that high-amplitude signals are associated with fast motion while while low-amplitude signals are associated with slow motion.
% We also add an explanation that the signal that has high variance from negative to positive values is associated with fast motion, whereas the signal stay closer to the zero is associated with rest motion.
The interface explained the model's purpose and accuracy, which was fixed to 90\% to remove potential bias.
For each model, the interface presents prediction results in two different conditions: (baseline) input and result, and (confidence score) input, result, and \textsc{confidence score}.
For each result, we asked participants how much they trust the model's prediction and whether they would be willing to make a medical decision based on that result.
%\cj{maybe a series of screenshots of user study?}
Participants saw a total of 24 scenarios (3 data types (image, audio, motion) $\times$ 2 conditions (baseline vs. \textsc{confidence score} $\times$ 2 \textsc{confidence score} (high vs. low) $\times$ 2 results (positive vs. negative).
To provide realistic experience, we provide different skin tone images for skin lesion samples based on the reported skin tone.
With the exception of the data type, the scenarios were shuffled across all other factors to avoid any ordering effects.
After participants saw all of the scenarios for a given data type, we presented them with five data examples and asked them to pick the ones that the model would be confident in processing according to \textsc{confidence score}.
We added these questions to assess whether people were able to learn about how the \textsc{confidence score} was being generated after seeing a series of examples.

\begin{figure}[t]
    \centering
    \includegraphics[width=1.0\linewidth]{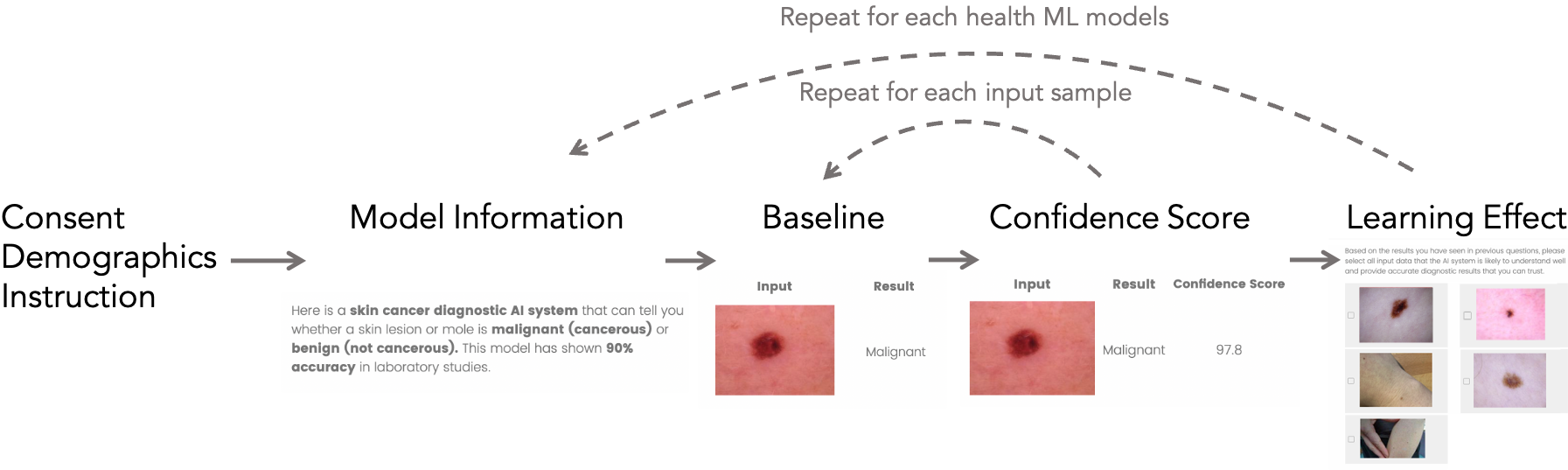}
    \caption{User study overview. The participants are first asked to give consent, read instruction, and provide demographics. Then, they report perceived trustworthiness and willingness to make a medical decision after seeing input samples that consist of different data types, diagnostic results, and \textsc{confidence score} for baseline and \textsc{confidence score} condition. Screenshots of the user study interface are demonstrated in Appendix~\ref{sec:user-study-interface}.}
    \label{fig:user-study}
\end{figure}

\revise{We intend to target ordinary, non-expert consumers at random rather than expert clinicians for our study. Our research is primarily directed toward the boom in consumer-facing mobile health applications, where non-expert users are expected to collect input data themselves. We believe this is where models are most susceptible to out-of-distribution inputs, providing unreliable predictions to the users. For the safe use of AI-powered health applications, the users would need support via automated uncertainty measures.
To this end}, We recruited participants from Amazon Mechanical Turk and compensated with \$3 USD for a 15-min online study. In total, 192 participants (155 male, 67 female) completed the online study with an average age of 42.7 $\pm$ 9.1 years.
The study was approved by Institutional Review Boards at the University of Washington. 

\begin{table}[]
\centering
\resizebox{\textwidth}{!}{
\begin{tabular}{rrrrrlr}\toprule
 & \multicolumn{3}{c}{\textbf{User-perceived trustworthiness}} & \multicolumn{3}{c}{\textbf{Impact on making medical decisions}} \\
% & & & Rank-Biserial & & & Rank-Biserial \\
 & Wilcoxon Test ($W$) & $p$ & Effect Size ($r$) & Wilcoxon Test ($W$) & \multicolumn{1}{r}{$p$} & Effect Size ($r$) \\\midrule
All & 529,950.0 & $<0.001^{***}$ & 0.393 & 223,227.0 & $<0.001^{***}$ & 0.178 \\
In-Distribution & 138,778.5 & $<0.001^{***}$ & 0.475 & 52,056.0 & $<0.001^{***}$ & 0.200 \\
Out-of-distribution & 126,814.0 & $<0.001^{***}$ & 0.317 & 59,790.5 & $0.001^{***}$ & 0.158 \\
Negative result & 131,910.0 & $<0.001^{***}$ & 0.393 & 51,197.5 & $0.026^{*}$ & 0.100 \\
Positive result & 133,301.0 & $<0.001^{***}$ & 0.394 & 60,751.5 & $<0.001^{***}$ & 0.258 \\
Image & 55,890.0 & $<0.001^{***}$ & 0.436 & 20,884.0 & $<0.001^{***}$ & 0.225 \\
Audio & 64,440.0 & $<0.001^{***}$ & 0.384 & 25,848.0 & $0.002^{**}$ & 0.173 \\
Motion & 56,767.5 & $<0.001^{***}$ & 0.361 & 28,084.5 & $0.019^{*}$ & 0.133 \\\bottomrule
\end{tabular}}
\caption{Results of Wilcoxon test in comparing baseline and \textsc{confidence score} conditions for the perceived trustworthiness and impact on decision making. All comparisons show statistically significant results. (***: $p < 0.001$, **: $p < 0.01$, *: $p < 0.05$).}
\label{tab:user-study}
\end{table}

\subsection{Results}
We analyzed the responses using the Wilcoxon signed-rank test~\cite{wilcoxon1992individual} to compute a pairwise comparison of the categorical responses between the baseline and \textsc{confidence score} conditions.
Table~\ref{tab:user-study} summarizes these statistical results along with the Rosenthal correlation coefficient~\cite{rosenthal1991meta} ($r$) for effect size. %\cj{add something about we want users to trust the results with high confidence score and not to trust the results with low confidence score.}

\textbf{RQ.~1: User-Perceived Trustworthiness} In general, we found that user-perceived trustworthiness ($p<0.001$) was higher in the \textsc{confidence score} condition with medium effect size ($r=0.393$). 
In other words, the dataset shift information helped the participants decide when to trust or not to trust the output of the models. 
Higher \textsc{confidence scores} led to increasing trustworthiness; high scores had a large effect size ($r=0.475$), while low scores had a medium effect size ($r=0.317$).
%This implies that the dataset shift information can be used to enhance the trustworthiness of the diagnostic results
The impact on trustworthiness was similar for positive and negative diagnostic results.
The effect sizes varied for the different input data types, with the images having the largest effect size and motion having the smallest.
We suspect that the effect size was correlated with the intuitiveness of the data types, with images being more intuitive than motion data.
%For example, the image input data are easily interpretable by the users and their trustworthiness was more affected by \textsc{confidence score} presented along with the input data. \cj{poor wording. rewrite}

\textbf{RQ.~2: Impact on Making Medical Decisions} 
When we examined the impact of \textsc{confidence score} on making medical decisions, we found that there was a statistically significant difference ($p<0.001$) between the baseline and \textsc{confidence score} conditions.
In other words, participants were more willing to make medical decisions when positive results were presented with high \textsc{confidence score} and vice versa.
Similar to the results for user-perceived trustworthiness, the effect of \textsc{confidence score} was larger on input data with high scores than low scores and highest for images compared to audio and motion data.

\textbf{RQ.~3: Learning Effect on Distinguishing In- and Out-of-Distribution Input Data} 
We found that the participants were able to learn from their interaction with \textsc{confidence score}.
The average Jaccard index when it came to selecting high \textsc{confidence score} input data was 0.75, 0.66, and 0.64 for image, audio, and motion data, respectively, which is a moderately high similarity. 
%This shows that the participants can distinguish between in- and out-of-distribution input data with a moderate accuracy.
As with our other results, the Jaccard index was highest on images and lowest on the motion data, implying that ability to understand the input data also has impact on learning effect.
%The learning effect implies that the dataset shift information can be used to guide health application users to capture the input data that the machine learning models can interpret. 
This implies that the dataset shift information can make users better understand input data that the machine learning models can interpret for the future interaction. 
% Furthermore, as the users interact with the models with \textsc{confidence score} longer, they can accurately discern whether the captured input data would be appropriate for the models, preventing the models from providing unsafe, unreasonable diagnostic results.

\vspace{-0.2cm}
\section{Discussion} 
\vspace{-0.2cm}
\label{sec:discussion}
\textbf{Dataset Shift Information for Health Application Users} Based on the results from our performance evaluation and user study, we can imagine two potential use cases of the dataset shift information to improve safety and trust in mHealth applications.
First, mHealth applications with machine learning models can exclude out-of-distribution samples to avoid making inferences and suggestions that are likely to be inaccurate and unreasonable.
%The app can gracefully guide the users to capture the input if the input is determined to not belong to in-distribution dataset.
% This approach provides reliable results by disregarding uncertain predictions and assures users' safety since they will not be exposed to unreasonable diagnostic results.
Second, our user study shows that the dataset shift information can enhance their interaction with the health machine learning applications.
It was found to be particularly effective in improving trustworthiness for in-distribution data and leading the users to make the right medical decision.
As the users interact with the health applications longer, they would have better understanding of importance of data quality for future interactions.
%Along with the background filtering of out-of-distribution input, we recommend to show the dataset shift information for in-distribution data for the users to trust the diagnostic prediction results.
%We argue health machine learning applications should actively

\textbf{Dataset Shift Information for Health Application Developers} Our dataset shift information not only improves the user experience, but also yields potential benefits for model developers.
If a user correctly captures data but the model rejects it as being out-of-distribution, then there likely exists intrinsic problems or biases with the model.
For example, if a skin lesion classifier is only trained on data from people with pale skin and a user with darker skin submits an image of their own, the out-of-distribution detector be triggered due to the incompleteness of the training dataset.
% In the case of the skin lesion classifier, if the train set only includes skin lesion images for people of a certain skin tone, inputs from users with different skin tones can be detected as an out-of-distribution input.
% The information can be used to reduce the biases in the dataset.
The same issues may occur when training dataset is only collected from a specific set of sensors (e.g., camera, microphone, IMU) with particular specifications.
% When the users' input data is captured with different devices, the input data can be considered as out-of-distribution.
% By monitoring the failures with dataset shift information, developers can identify the potential bias or limitations of the models and the accompanying training dataset.

\textbf{Limitations and Future Work}
%Although the out-of-distribution detectors we evaluated in Section~\ref{sec:tech-study} show promising results, not all out-of-distribution works well on when evaluated outside of their own experiment setting; energy-based out-of-distribution detector did not show acceptable performance.
%While this work did not further investigate the cause of its poor performance, this could be an interesting future work to explore.
%In the same context, as out-of-distribution detection is actively investigated and constantly improved, the benefits of using dataset shift information would be also magnified.
Detecting near-distribution samples (e.g., ISIC 2017, Stethoscope, Kaggle Parkinson's) was a difficult problem for all the out-of-distribution detectors we evaluated. 
%Furthermore, we did not find a meaningful impact of out-of-distribution detection on the classification accuracy.
For the near-distribution datasets, we evaluated model's accuracy on the data that are distinguished as in-distribution. 
This issue is actively investigated by researchers and the improved near-distribution detection method would benefit this work.
%There was not a significant classification accuracy improvement (e.g., 3\%, 5\% improvement for ISIC2017 and ISIC2016).
%This implies that even when some input data are detected as in-distribution, it does not guarantee to improve classification accuracy. 

Our user study was limited in the fact that it dealt with hypothetical scenarios.
There were no repercussions for users decisions, so they may not have spent as much time making their decisions as they would in real life.
There are also many other factors that impact people's health-related decision making, such as the perceived severity of the medical condition and the perceived benefits of taking action~\cite{janz1984health, strecher1997health}.
We tried to make some of the data more realistic by aligning data with the user's demographic information (e.g., we displayed skin lesion images based on their reported skin tone); nevertheless, participants were aware that the data was not their own.
\revise{Additionally, increased trust might be affected by participants' own understanding and interpretation of the input data. Although we observed increased and decreased trust in examples with high and low \textsc{confidence score}, respectively, randomizing the \textsc{confidence score} of input data could further quantify impact of \textsc{confidence score} on user trust of health predictions.}
% For the user study, the example input data shown to the participants could be different from their own skin, voice, or motion.
% This disparity could make the participants difficult to fully engage in the study, potentially impacting their responses.
% Additionally, the gender and age of the participants are not equally balanced; there are more male participants in their 30's.
% This implies that the user study results are not generalizeable to all demographics.

% As this work shows promising results on using dataset shift information for more reliable and safer interaction with health machine learning applications.
In future work, we would like to (1) investigate the best way to present this information for the users, (2) leverage the dataset shift information for finding potential biases in the train dataset and inherent problems with the model, (3) investigate an out-of-distribution method for better near-distribution detection performance. 

%\subsection{Compatibility with Other Out-of-Distribution Detectors}

%\subsection{Potential Use of the Confidence Score}

%\subsection{Bias and Data Quality}

%\subsection{}
\vspace{-0.2cm}
\section{Conclusion}
\vspace{-0.2cm}
In this work, we investigated the utility of dataset shift information for improving reliability and trustworthiness of machine learning-based health applications.
Using publicly available health deep learning models and datasets, we first demonstrated that the models fail when encountered with unseen data.
We then evaluated the out-of-distribution detection performance of state-of-the-art methods, showing high accuracy in distinguishing between in- and out-of-distribution datasets for different input data types (e.g., image, audio, motion data).
We conducted an online user study to investigate the effect of dataset shift information on potential users.
We found that the participants trusted prediction results with high \textsc{confidence score} and are more willing to make a right medical decision, while they considered prediction results with low \textsc{confidence score} less trustworthy and are less willing to make medical decision.
This work shows that the dataset shift is a meaningful piece of information for building consumer-facing trustworthy AI applications for high-stake decision making.
\section{Acknowledgments and Disclosure of Funding}
\revise{We would like to thank Alex Mariakakis for insightful feedback and discussions.
This work was supported by NSF CNS-1565252. Shwetak Patel is on partial leave at Google.}

{\small
\bibliography{references}}

\begin{thebibliography}{10}

\bibitem{abadi2016deep}
Martin {Abadi}, Andy {Chu}, Ian {Goodfellow}, H.~Brendan {McMahan}, Ilya
  {Mironov}, Kunal {Talwar}, and Li~{Zhang}.
\newblock Deep learning with differential privacy.
\newblock In {\em Proceedings of the 2016 ACM SIGSAC Conference on Computer and
  Communications Security}, pages 308--318, 2016.

\bibitem{adebayo2018sanity}
Julius {Adebayo}, Justin {Gilmer}, Michael~Christoph {Muelly}, Ian
  {Goodfellow}, Moritz {Hardt}, and Been {Kim}.
\newblock Sanity checks for saliency maps.
\newblock In {\em Advances in Neural Information Processing Systems},
  volume~31, pages 9505--9515, 2018.

\bibitem{agarwal2019fair}
Alekh {Agarwal}, Miroslav {Dudík}, and Zhiwei~Steven {Wu}.
\newblock Fair regression: Quantitative definitions and reduction-based
  algorithms.
\newblock In {\em 36th International Conference on Machine Learning, ICML
  2019}, pages 120--129, 2019.

\bibitem{akbar2020safety}
Saba Akbar, Enrico Coiera, and Farah Magrabi.
\newblock Safety concerns with consumer-facing mobile health applications and
  their consequences: a scoping review.
\newblock {\em Journal of the American Medical Informatics Association},
  27(2):330--340, 2020.

\bibitem{ardila2019end}
Diego Ardila, Atilla~P Kiraly, Sujeeth Bharadwaj, Bokyung Choi, Joshua~J
  Reicher, Lily Peng, Daniel Tse, Mozziyar Etemadi, Wenxing Ye, Greg Corrado,
  et~al.
\newblock End-to-end lung cancer screening with three-dimensional deep learning
  on low-dose chest computed tomography.
\newblock {\em Nature medicine}, 25(6):954--961, 2019.

\bibitem{athey2017beyond}
Susan {Athey}.
\newblock Beyond prediction: Using big data for policy problems.
\newblock {\em Science}, 355(6324):483--485, 2017.

\bibitem{banos2014mhealthdroid}
Oresti Banos, Rafael Garcia, Juan~A Holgado-Terriza, Miguel Damas, Hector
  Pomares, Ignacio Rojas, Alejandro Saez, and Claudia Villalonga.
\newblock mhealthdroid: a novel framework for agile development of mobile
  health applications.
\newblock In {\em International workshop on ambient assisted living}, pages
  91--98. Springer, 2014.

\bibitem{bhatt2020uncertainty}
Umang Bhatt, Javier Antor{\'a}n, Yunfeng Zhang, Q~Vera Liao, Prasanna
  Sattigeri, Riccardo Fogliato, Gabrielle~Gauthier Melan{\c{c}}on, Ranganath
  Krishnan, Jason Stanley, Omesh Tickoo, et~al.
\newblock Uncertainty as a form of transparency: Measuring, communicating, and
  using uncertainty.
\newblock {\em arXiv preprint arXiv:2011.07586}, 2020.

\bibitem{bot2016mpower}
Brian~M Bot, Christine Suver, Elias~Chaibub Neto, Michael Kellen, Arno Klein,
  Christopher Bare, Megan Doerr, Abhishek Pratap, John Wilbanks, E~Ray Dorsey,
  et~al.
\newblock The mpower study, parkinson disease mobile data collected using
  researchkit.
\newblock {\em Scientific data}, 3(1):1--9, 2016.

\bibitem{ccalli2019frodo}
Erdi {\c{C}}all{\i}, Keelin Murphy, Ecem Sogancioglu, and Bram Van~Ginneken.
\newblock Frodo: Free rejection of out-of-distribution samples: application to
  chest x-ray analysis.
\newblock {\em arXiv preprint arXiv:1907.01253}, 2019.

\bibitem{cao2020benchmark}
Tianshi Cao, Chinwei Huang, David Yu-Tung Hui, and Joseph~Paul Cohen.
\newblock A benchmark of medical out of distribution detection.
\newblock {\em arXiv preprint arXiv:2007.04250}, 2020.

\bibitem{codella2019skin}
Noel Codella, Veronica Rotemberg, Philipp Tschandl, M~Emre Celebi, Stephen
  Dusza, David Gutman, Brian Helba, Aadi Kalloo, Konstantinos Liopyris, Michael
  Marchetti, et~al.
\newblock Skin lesion analysis toward melanoma detection 2018: A challenge
  hosted by the international skin imaging collaboration (isic).
\newblock {\em arXiv preprint arXiv:1902.03368}, 2019.

\bibitem{codella2018skin}
Noel~CF Codella, David Gutman, M~Emre Celebi, Brian Helba, Michael~A Marchetti,
  Stephen~W Dusza, Aadi Kalloo, Konstantinos Liopyris, Nabin Mishra, Harald
  Kittler, et~al.
\newblock Skin lesion analysis toward melanoma detection: A challenge at the
  2017 international symposium on biomedical imaging (isbi), hosted by the
  international skin imaging collaboration (isic).
\newblock In {\em 2018 IEEE 15th International Symposium on Biomedical Imaging
  (ISBI 2018)}, pages 168--172. IEEE, 2018.

\bibitem{de2018clinically}
Jeffrey De~Fauw, Joseph~R Ledsam, Bernardino Romera-Paredes, Stanislav Nikolov,
  Nenad Tomasev, Sam Blackwell, Harry Askham, Xavier Glorot, Brendan
  O’Donoghue, Daniel Visentin, et~al.
\newblock Clinically applicable deep learning for diagnosis and referral in
  retinal disease.
\newblock {\em Nature medicine}, 24(9):1342--1350, 2018.

\bibitem{de2014bilicam}
Lilian De~Greef, Mayank Goel, Min~Joon Seo, Eric~C Larson, James~W Stout,
  James~A Taylor, and Shwetak~N Patel.
\newblock Bilicam: using mobile phones to monitor newborn jaundice.
\newblock In {\em Proceedings of the 2014 ACM International Joint Conference on
  Pervasive and Ubiquitous Computing}, pages 331--342, 2014.

\bibitem{DeBruine2021}
Lisa DeBruine and Benedict Jones.
\newblock {Face Research Lab London Set}.
\newblock 4 2021.

\bibitem{dudik2020assessing}
Miro {Dudík}, William {Chen}, Solon {Barocas}, Mario {Inchiosa}, Nick
  {Lewins}, Miruna {Oprescu}, Joy {Qiao}, Mehrnoosh {Sameki}, Mario {Schlener},
  Jason {Tuo}, and Hanna {Wallach}.
\newblock Assessing and mitigating unfairness in credit models with the
  fairlearn toolkit.
\newblock 2020.

\bibitem{evtimov2020security}
Ivan {Evtimov}, Weidong {Cui}, Ece {Kamar}, Emre {Kiciman}, Tadayoshi {Kohno},
  and Jerry {Li}.
\newblock Security and machine learning in the real world.
\newblock {\em arXiv preprint arXiv:2007.07205}, 2020.

\bibitem{eykholt2018robust}
Kevin {Eykholt}, Ivan {Evtimov}, Earlence {Fernandes}, Bo~{Li}, Amir {Rahmati},
  Chaowei {Xiao}, Atul {Prakash}, Tadayoshi {Kohno}, and Dawn {Song}.
\newblock Robust physical-world attacks on deep learning visual classification.
\newblock In {\em 2018 IEEE/CVF Conference on Computer Vision and Pattern
  Recognition}, pages 1625--1634, 2018.

\bibitem{felzmann2019transparency}
Heike {Felzmann}, Eduard~Fosch {Villaronga}, Christoph {Lutz}, and Aurelia
  {Tamò-Larrieux}.
\newblock Transparency you can trust: Transparency requirements for artificial
  intelligence between legal norms and contextual concerns:.
\newblock {\em Big Data \& Society}, 6(1):1--14, 2019.

\bibitem{ford2019adversarial}
Nic {Ford}, Justin {Gilmer}, Nicolas {Carlini}, and Dogus {Cubuk}.
\newblock Adversarial examples are a natural consequence of test error in
  noise.
\newblock {\em arXiv preprint arXiv:1901.10513}, 2019.

\bibitem{stetho2021}
Mohammad Fraiwan, Luay Fraiwan, Basheer Khassawneh, and Ali Ibnian.
\newblock A dataset of lung sounds recorded from the chest wall using an
  electronic stethoscope.
\newblock {\em Data in Brief}, 35:106913, 2021.

\bibitem{gairola2020respirenet}
Siddhartha Gairola, Francis Tom, Nipun Kwatra, and Mohit Jain.
\newblock Respirenet: A deep neural network for accurately detecting abnormal
  lung sounds in limited data setting, 2020.

\bibitem{audioset2017}
Jort~F. Gemmeke, Daniel P.~W. Ellis, Dylan Freedman, Aren Jansen, Wade
  Lawrence, R.~Channing Moore, Manoj Plakal, and Marvin Ritter.
\newblock Audio set: An ontology and human-labeled dataset for audio events.
\newblock In {\em Proc. IEEE ICASSP 2017}, New Orleans, LA, 2017.

\bibitem{giorgia_2018}
Giorgia.
\newblock Simulation of parkinson movement disorders -- kaggle, May 2018.

\bibitem{gkotsis2017characterisation}
George Gkotsis, Anika Oellrich, Sumithra Velupillai, Maria Liakata, Tim~JP
  Hubbard, Richard~JB Dobson, and Rina Dutta.
\newblock Characterisation of mental health conditions in social media using
  informed deep learning.
\newblock {\em Scientific reports}, 7(1):1--11, 2017.

\bibitem{goodfellow2014explaining}
Ian~J Goodfellow, Jonathon Shlens, and Christian Szegedy.
\newblock Explaining and harnessing adversarial examples.
\newblock {\em arXiv preprint arXiv:1412.6572}, 2014.

\bibitem{hendrycks2019benchmarking}
Dan {Hendrycks} and Thomas~G. {Dietterich}.
\newblock Benchmarking neural network robustness to common corruptions and
  perturbations.
\newblock In {\em International Conference on Learning Representations}, 2019.

\bibitem{hendrycks2016a}
Dan {Hendrycks} and Kevin {Gimpel}.
\newblock A baseline for detecting misclassified and out-of-distribution
  examples in neural networks.
\newblock In {\em ICLR (Poster)}, 2016.

\bibitem{hendrycks2018deep}
Dan Hendrycks, Mantas Mazeika, and Thomas Dietterich.
\newblock Deep anomaly detection with outlier exposure.
\newblock {\em arXiv preprint arXiv:1812.04606}, 2018.

\bibitem{hoffman2021smartphone}
Jason~S Hoffman, Varun Viswanath, Xinyi Ding, Matthew~J Thompson, Eric~C
  Larson, Shwetak~N Patel, and Edward Wang.
\newblock Smartphone camera oximetry in an induced hypoxemia study.
\newblock {\em arXiv preprint arXiv:2104.00038}, 2021.

\bibitem{huynh2019vitamon}
Sinh Huynh, Rajesh~Krishna Balan, JeongGil Ko, and Youngki Lee.
\newblock Vitamon: Measuring heart rate variability using smartphone front
  camera.
\newblock In {\em Proceedings of the 17th Conference on Embedded Networked
  Sensor Systems}, pages 1--14, 2019.

\bibitem{janz1984health}
Nancy~K Janz and Marshall~H Becker.
\newblock The health belief model: A decade later.
\newblock {\em Health education quarterly}, 11(1):1--47, 1984.

\bibitem{jiang2018to}
Heinrich {Jiang}, Been {Kim}, Melody~Y. {Guan}, and Maya~R. {Gupta}.
\newblock To trust or not to trust a classifier.
\newblock In {\em Advances in Neural Information Processing Systems},
  volume~31, pages 5541--5552, 2018.

\bibitem{kaur2020interpreting}
Harmanpreet {Kaur}, Harsha {Nori}, Samuel {Jenkins}, Rich {Caruana}, Hanna
  {Wallach}, and Jennifer~Wortman {Vaughan}.
\newblock Interpreting interpretability: Understanding data scientists' use of
  interpretability tools for machine learning.
\newblock In {\em Proceedings of the 2020 CHI Conference on Human Factors in
  Computing Systems}, pages 1--14, 2020.

\bibitem{kim2018interpretability}
Been Kim, Martin Wattenberg, Justin Gilmer, Carrie Cai, James Wexler, Fernanda
  Viegas, et~al.
\newblock Interpretability beyond feature attribution: Quantitative testing
  with concept activation vectors (tcav).
\newblock In {\em International conference on machine learning}, pages
  2668--2677. PMLR, 2018.

\bibitem{krizhevsky2009learning}
Alex Krizhevsky, Geoffrey Hinton, et~al.
\newblock Learning multiple layers of features from tiny images.
\newblock 2009.

\bibitem{lakshminarayanan2017simple}
Balaji {Lakshminarayanan}, Alexander {Pritzel}, and Charles {Blundell}.
\newblock Simple and scalable predictive uncertainty estimation using deep
  ensembles.
\newblock In {\em Advances in Neural Information Processing Systems},
  volume~30, pages 6402--6413, 2017.

\bibitem{lee2017training}
Kimin Lee, Honglak Lee, Kibok Lee, and Jinwoo Shin.
\newblock Training confidence-calibrated classifiers for detecting
  out-of-distribution samples.
\newblock {\em arXiv preprint arXiv:1711.09325}, 2017.

\bibitem{lee2018}
Kimin Lee, Kibok Lee, Honglak Lee, and Jinwoo Shin.
\newblock A simple unified framework for detecting out-of-distribution samples
  and adversarial attacks.
\newblock In S.~Bengio, H.~Wallach, H.~Larochelle, K.~Grauman, N.~Cesa-Bianchi,
  and R.~Garnett, editors, {\em Advances in Neural Information Processing
  Systems}, volume~31. Curran Associates, Inc., 2018.

\bibitem{liang2018enhancing}
Shiyu {Liang}, Yixuan {Li}, and Rayadurgam {Srikant}.
\newblock Enhancing the reliability of out-of-distribution image detection in
  neural networks.
\newblock In {\em 6th International Conference on Learning Representations,
  ICLR 2018}, 2018.

\bibitem{lipton2018detecting}
Zachary Lipton, Yu-Xiang Wang, and Alexander Smola.
\newblock Detecting and correcting for label shift with black box predictors.
\newblock In {\em International conference on machine learning}, pages
  3122--3130. PMLR, 2018.

\bibitem{energy2020}
Weitang Liu, Xiaoyun Wang, John Owens, and Yixuan Li.
\newblock Energy-based out-of-distribution detection.
\newblock In H.~Larochelle, M.~Ranzato, R.~Hadsell, M.~F. Balcan, and H.~Lin,
  editors, {\em Advances in Neural Information Processing Systems}, volume~33,
  pages 21464--21475. Curran Associates, Inc., 2020.

\bibitem{liu2020multi}
Xin Liu, Josh Fromm, Shwetak Patel, and Daniel McDuff.
\newblock Multi-task temporal shift attention networks for on-device
  contactless vitals measurement.
\newblock {\em arXiv preprint arXiv:2006.03790}, 2020.

\bibitem{lundberg2017a}
Scott~M. {Lundberg} and Su-In {Lee}.
\newblock A unified approach to interpreting model predictions.
\newblock In {\em NIPS'17 Proceedings of the 31st International Conference on
  Neural Information Processing Systems}, volume~30, pages 4768--4777, 2017.

\bibitem{motionsense2019}
Mohammad Malekzadeh, Richard~G. Clegg, Andrea Cavallaro, and Hamed Haddadi.
\newblock Mobile sensor data anonymization.
\newblock In {\em Proceedings of the International Conference on Internet of
  Things Design and Implementation}, IoTDI '19, pages 49--58, New York, NY,
  USA, 2019. ACM.

\bibitem{malinin2018predictive}
Andrey Malinin and Mark Gales.
\newblock Predictive uncertainty estimation via prior networks.
\newblock {\em arXiv preprint arXiv:1802.10501}, 2018.

\bibitem{mariakakis2017biliscreen}
Alex Mariakakis, Megan~A Banks, Lauren Phillipi, Lei Yu, James Taylor, and
  Shwetak~N Patel.
\newblock Biliscreen: smartphone-based scleral jaundice monitoring for liver
  and pancreatic disorders.
\newblock {\em Proceedings of the ACM on Interactive, Mobile, Wearable and
  Ubiquitous Technologies}, 1(2):1--26, 2017.

\bibitem{mariakakis2017pupilscreen}
Alex Mariakakis, Jacob Baudin, Eric Whitmire, Vardhman Mehta, Megan~A Banks,
  Anthony Law, Lynn Mcgrath, and Shwetak~N Patel.
\newblock Pupilscreen: using smartphones to assess traumatic brain injury.
\newblock {\em Proceedings of the ACM on Interactive, Mobile, Wearable and
  Ubiquitous Technologies}, 1(3):1--27, 2017.

\bibitem{Mariakakis2019}
Alex Mariakakis, Edward Wang, Shwetak Patel, and Mayank Goel.
\newblock Challenges in realizing smartphone-based health sensing.
\newblock {\em IEEE Pervasive Computing}, 18(2):76--84, apr 2019.

\bibitem{melis2019exploiting}
Luca {Melis}, Congzheng {Song}, Emiliano~De {Cristofaro}, and Vitaly
  {Shmatikov}.
\newblock Exploiting unintended feature leakage in collaborative learning.
\newblock In {\em 2019 IEEE Symposium on Security and Privacy (SP)}, pages
  691--706, 2019.

\bibitem{mohseni2020self}
Sina Mohseni, Mandar Pitale, JBS Yadawa, and Zhangyang Wang.
\newblock Self-supervised learning for generalizable out-of-distribution
  detection.
\newblock In {\em Proceedings of the AAAI Conference on Artificial
  Intelligence}, volume~34, pages 5216--5223, 2020.

\bibitem{morningstar2021density}
Warren Morningstar, Cusuh Ham, Andrew Gallagher, Balaji Lakshminarayanan, Alex
  Alemi, and Joshua Dillon.
\newblock Density of states estimation for out of distribution detection.
\newblock In {\em International Conference on Artificial Intelligence and
  Statistics}, pages 3232--3240. PMLR, 2021.

\bibitem{nalisnick2019detecting}
Eric Nalisnick, Akihiro Matsukawa, Yee~Whye Teh, and Balaji Lakshminarayanan.
\newblock Detecting out-of-distribution inputs to deep generative models using
  a test for typicality.
\newblock {\em arXiv preprint arXiv:1906.02994}, 5:5, 2019.

\bibitem{nie2020quasi}
Xinkun {Nie} and Stefan {Wager}.
\newblock Quasi-oracle estimation of heterogeneous treatment effects.
\newblock {\em Biometrika}, 2020.

\bibitem{pacheco2020out}
Andre~GC Pacheco, Chandramouli~S Sastry, Thomas Trappenberg, Sageev Oore, and
  Renato~A Krohling.
\newblock On out-of-distribution detection algorithms with deep neural skin
  cancer classifiers.
\newblock In {\em Proceedings of the IEEE/CVF Conference on Computer Vision and
  Pattern Recognition Workshops}, pages 732--733, 2020.

\bibitem{papernot2018scalable}
Nicolas {Papernot}, Shuang {Song}, Ilya {Mironov}, Ananth {Raghunathan}, Kunal
  {Talwar}, and Úlfar {Erlingsson}.
\newblock Scalable private learning with pate.
\newblock In {\em International Conference on Learning Representations}, 2018.

\bibitem{quintanilha2018detecting}
Igor~M. {Quintanilha}, Roberto de~M.~E.~{Filho}, José {Lezama}, Mauricio
  {Delbracio}, and Leonardo~O. {Nunes}.
\newblock Detecting out-of-distribution samples using low-order deep features
  statistics.
\newblock 2018.

\bibitem{raghavan2020mitigating}
Manish {Raghavan}, Solon {Barocas}, Jon {Kleinberg}, and Karen {Levy}.
\newblock Mitigating bias in algorithmic hiring: evaluating claims and
  practices.
\newblock In {\em Proceedings of the 2020 Conference on Fairness,
  Accountability, and Transparency}, pages 469--481, 2020.

\bibitem{ren2019likelihood}
Jie Ren, Peter~J Liu, Emily Fertig, Jasper Snoek, Ryan Poplin, Mark~A DePristo,
  Joshua~V Dillon, and Balaji Lakshminarayanan.
\newblock Likelihood ratios for out-of-distribution detection.
\newblock {\em arXiv preprint arXiv:1906.02845}, 2019.

\bibitem{rocha2019open}
Bruno~M Rocha, Dimitris Filos, Lu{\'\i}s Mendes, Gorkem Serbes, Sezer Ulukaya,
  Yasemin~P Kahya, Nik{\v{s}}a Jakovljevic, Tatjana~L Turukalo, Ioannis~M
  Vogiatzis, Eleni Perantoni, et~al.
\newblock An open access database for the evaluation of respiratory sound
  classification algorithms.
\newblock {\em Physiological measurement}, 40(3):035001, 2019.

\bibitem{rosenthal1991meta}
R.~Rosenthal, H.~Rosenthal, and inc Sage~Publications.
\newblock {\em Meta-Analytic Procedures for Social Research}.
\newblock Applied Social Research Methods. SAGE Publications, 1991.

\bibitem{roy2021does}
Abhijit~Guha Roy, Jie Ren, Shekoofeh Azizi, Aaron Loh, Vivek Natarajan, Basil
  Mustafa, Nick Pawlowski, Jan Freyberg, Yuan Liu, Zach Beaver, et~al.
\newblock Does your dermatology classifier know what it doesn't know? detecting
  the long-tail of unseen conditions.
\newblock {\em arXiv preprint arXiv:2104.03829}, 2021.

\bibitem{sastry2020detecting}
Chandramouli~Shama {Sastry} and Sageev {Oore}.
\newblock Detecting out-of-distribution examples with gram matrices.
\newblock In {\em ICML 2020: 37th International Conference on Machine
  Learning}, volume~1, pages 8491--8501, 2020.

\bibitem{schoettker2020blood}
Patrick Schoettker, Jean Degott, Gregory Hofmann, Martin Proen{\c{c}}a,
  Guillaume Bonnier, Alia Lemkaddem, Mathieu Lemay, Raoul Schorer, Urvan
  Christen, Jean-Fran{\c{c}}ois Knebel, et~al.
\newblock Blood pressure measurements with the optibp smartphone app validated
  against reference auscultatory measurements.
\newblock {\em Scientific Reports}, 10(1):1--12, 2020.

\bibitem{sieberts2021crowdsourcing}
Solveig~K Sieberts, Jennifer Schaff, Marlena Duda, B{\'a}lint~{\'A}rmin Pataki,
  Ming Sun, Phil Snyder, Jean-Francois Daneault, Federico Parisi, Gianluca
  Costante, Udi Rubin, et~al.
\newblock Crowdsourcing digital health measures to predict parkinson’s
  disease severity: the parkinson’s disease digital biomarker dream
  challenge.
\newblock {\em NPJ digital medicine}, 4(1):1--12, 2021.

\bibitem{strecher1997health}
Victor~J Strecher and Irwin~M Rosenstock.
\newblock The health belief model.
\newblock {\em Cambridge handbook of psychology, health and medicine}, 113:117,
  1997.

\bibitem{su2020deep}
Chang Su, Zhenxing Xu, Jyotishman Pathak, and Fei Wang.
\newblock Deep learning in mental health outcome research: a scoping review.
\newblock {\em Translational Psychiatry}, 10(1):1--26, 2020.

\bibitem{subbaswamy2020development}
Adarsh Subbaswamy and Suchi Saria.
\newblock From development to deployment: dataset shift, causality, and
  shift-stable models in health ai.
\newblock {\em Biostatistics}, 21(2):345--352, 2020.

\bibitem{thiagarajan2020calibrating}
Jayaraman~J Thiagarajan, Prasanna Sattigeri, Deepta Rajan, and Bindya
  Venkatesh.
\newblock Calibrating healthcare ai: Towards reliable and interpretable deep
  predictive models.
\newblock {\em arXiv preprint arXiv:2004.14480}, 2020.

\bibitem{tramer2021differentially}
Florian {Tramer} and Dan {Boneh}.
\newblock Differentially private learning needs better features (or much more
  data).
\newblock In {\em ICLR 2021: The Ninth International Conference on Learning
  Representations}, 2021.

\bibitem{tramer2018ensemble}
Florian {Tramèr}, Alexey {Kurakin}, Nicolas {Papernot}, Ian~J. {Goodfellow},
  Dan {Boneh}, and Patrick~D. {McDaniel}.
\newblock Ensemble adversarial training: Attacks and defenses.
\newblock In {\em 6th International Conference on Learning Representations,
  ICLR 2018}, 2018.

\bibitem{tschandl2018ham10000}
Philipp Tschandl, Cliff Rosendahl, and Harald Kittler.
\newblock The ham10000 dataset, a large collection of multi-source
  dermatoscopic images of common pigmented skin lesions.
\newblock {\em Scientific data}, 5(1):1--9, 2018.

\bibitem{venkatakrishnan2020self}
Abinav~Ravi Venkatakrishnan, Seong~Tae Kim, Rami Eisawy, Franz Pfister, and
  Nassir Navab.
\newblock Self-supervised out-of-distribution detection in brain ct scans.
\newblock {\em arXiv preprint arXiv:2011.05428}, 2020.

\bibitem{wilcoxon1992individual}
Frank Wilcoxon.
\newblock Individual comparisons by ranking methods.
\newblock In {\em Breakthroughs in statistics}, pages 196--202. Springer, 1992.

\bibitem{zhang2020deep}
Hanrui Zhang, Kaiwen Deng, Hongyang Li, Roger~L Albin, and Yuanfang Guan.
\newblock Deep learning identifies digital biomarkers for self-reported
  parkinson's disease.
\newblock {\em Patterns}, 1(3):100042, 2020.

\bibitem{zhang2020hybrid}
Hongjie Zhang, Ang Li, Jie Guo, and Yanwen Guo.
\newblock Hybrid models for open set recognition.
\newblock In {\em European Conference on Computer Vision}, pages 102--117.
  Springer, 2020.

\bibitem{zhang2013domain}
Kun Zhang, Bernhard Sch{\"o}lkopf, Krikamol Muandet, and Zhikun Wang.
\newblock Domain adaptation under target and conditional shift.
\newblock In {\em International Conference on Machine Learning}, pages
  819--827. PMLR, 2013.

\end{thebibliography}
\bibliographystyle{plain}
\newpage
\appendix

%\section{Dataset Preview}
%\cj{collage of images, spectrograms, time-series plot for each dataset}

%\newpage
\section{User Study Interface}
\label{sec:user-study-interface}
In this section, we provide screenshots and list of examples that were used in the user study.

\begin{figure}[h!]
    \centering
    \includegraphics[width=0.8\linewidth]{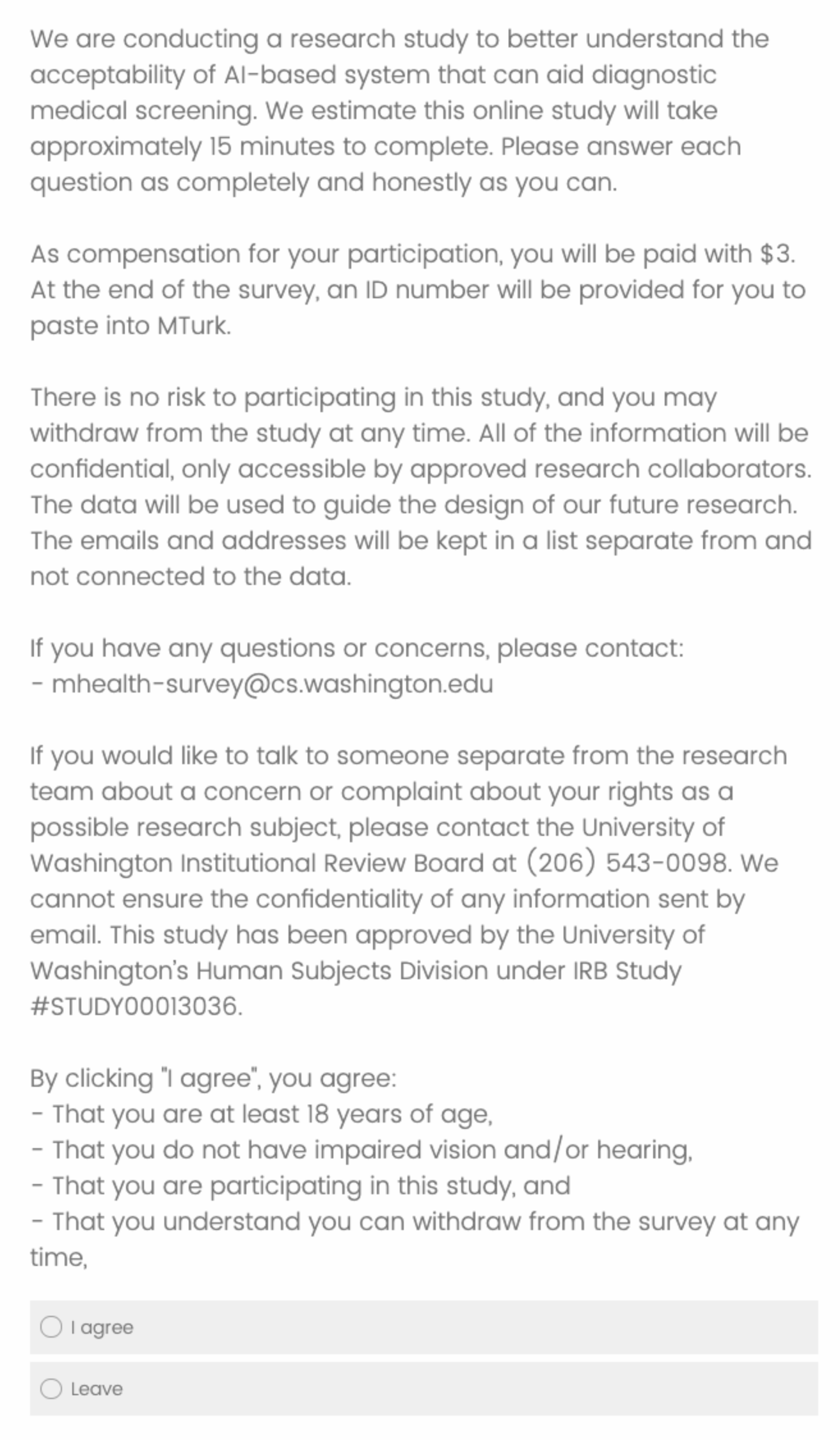}
    \caption{User study consent form. Note that the name of the institution is redacted for the review.}
    \label{fig:my_label}
\end{figure}

\newpage

\begin{figure}[h!]
    \centering
    \hfill\begin{subfigure}{0.38\linewidth}
    \vspace{-5.50cm}
    \includegraphics[width=\linewidth]{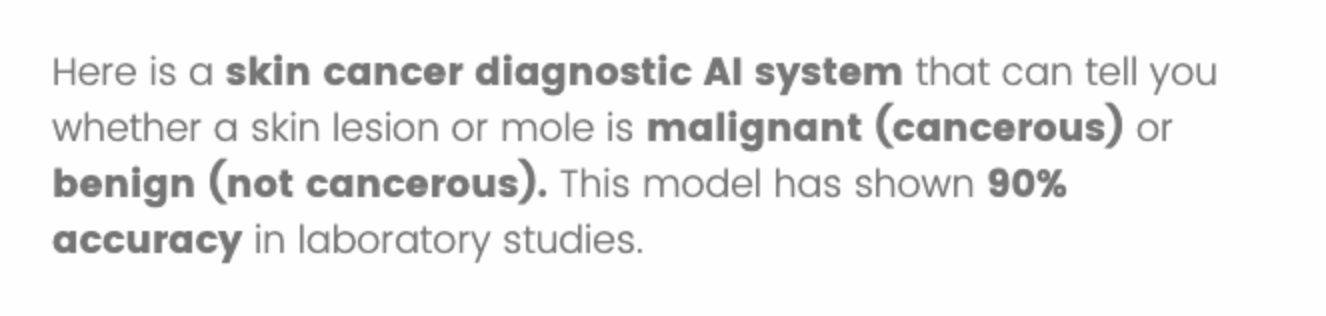}
    \caption{Interface that shows information about a health machine learning model. It shows target health condition, possible prediction results, and its accuracy. } \label{fig:interface-a}
    \end{subfigure}\hfill
    \begin{subfigure}{0.38\linewidth}
    \includegraphics[width=\linewidth]{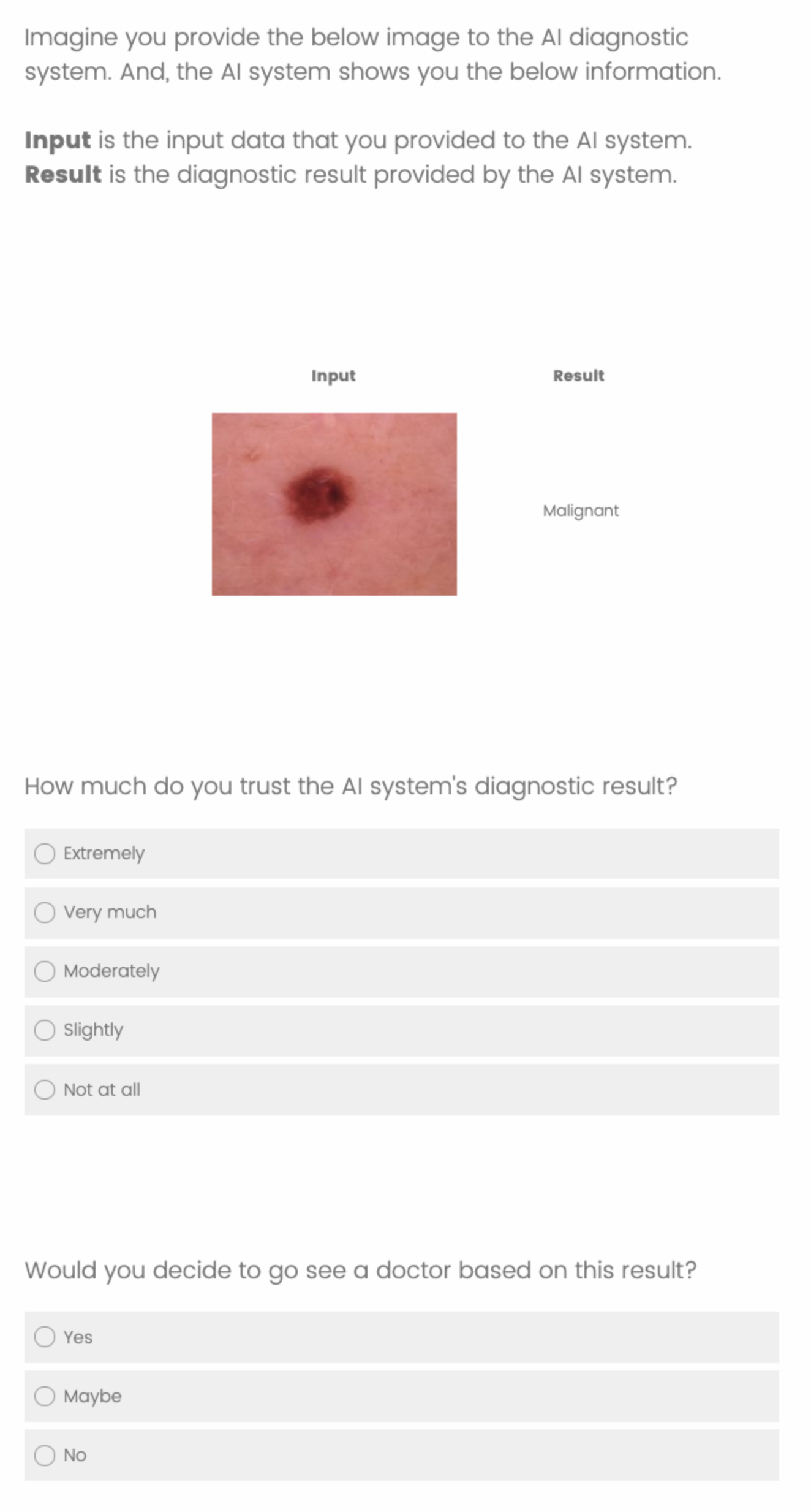}
    \caption{Interface that shows baseline condition. This condition only presents input data and prediction results and asks questions on user-perceived trustworthiness and impact on making medical decisions.} \label{fig:interface-b}
    \end{subfigure}\hfill
    
    \medskip
    
    \hfill\begin{subfigure}{0.38\linewidth}
    \vspace{-5.60cm}
    \includegraphics[width=\linewidth]{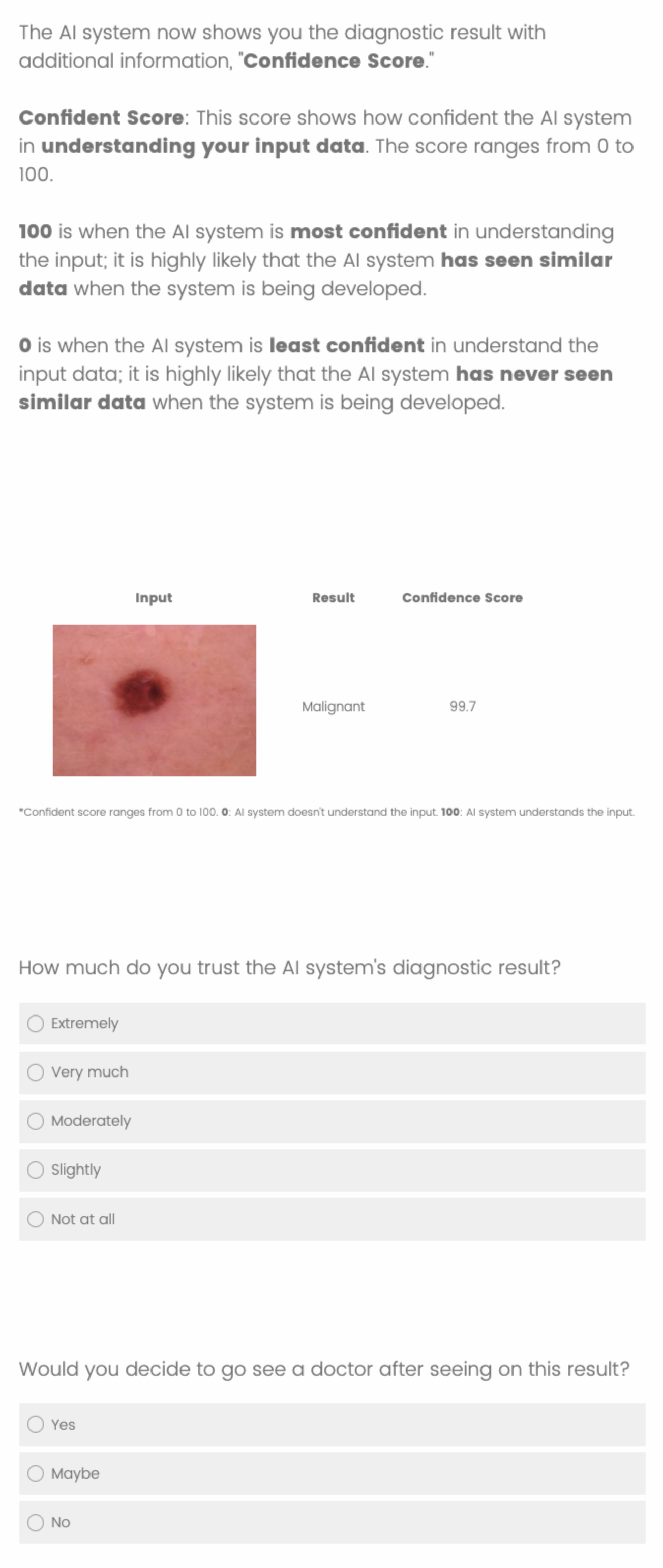}
    \caption{Interface that shows \textsc{confidence score} condition. This condition only presents input data, prediction results, and \textsc{confidence score}.} \label{fig:interface-c}
    \end{subfigure}\hfill
    \begin{subfigure}{0.38\linewidth}
    \includegraphics[width=\linewidth]{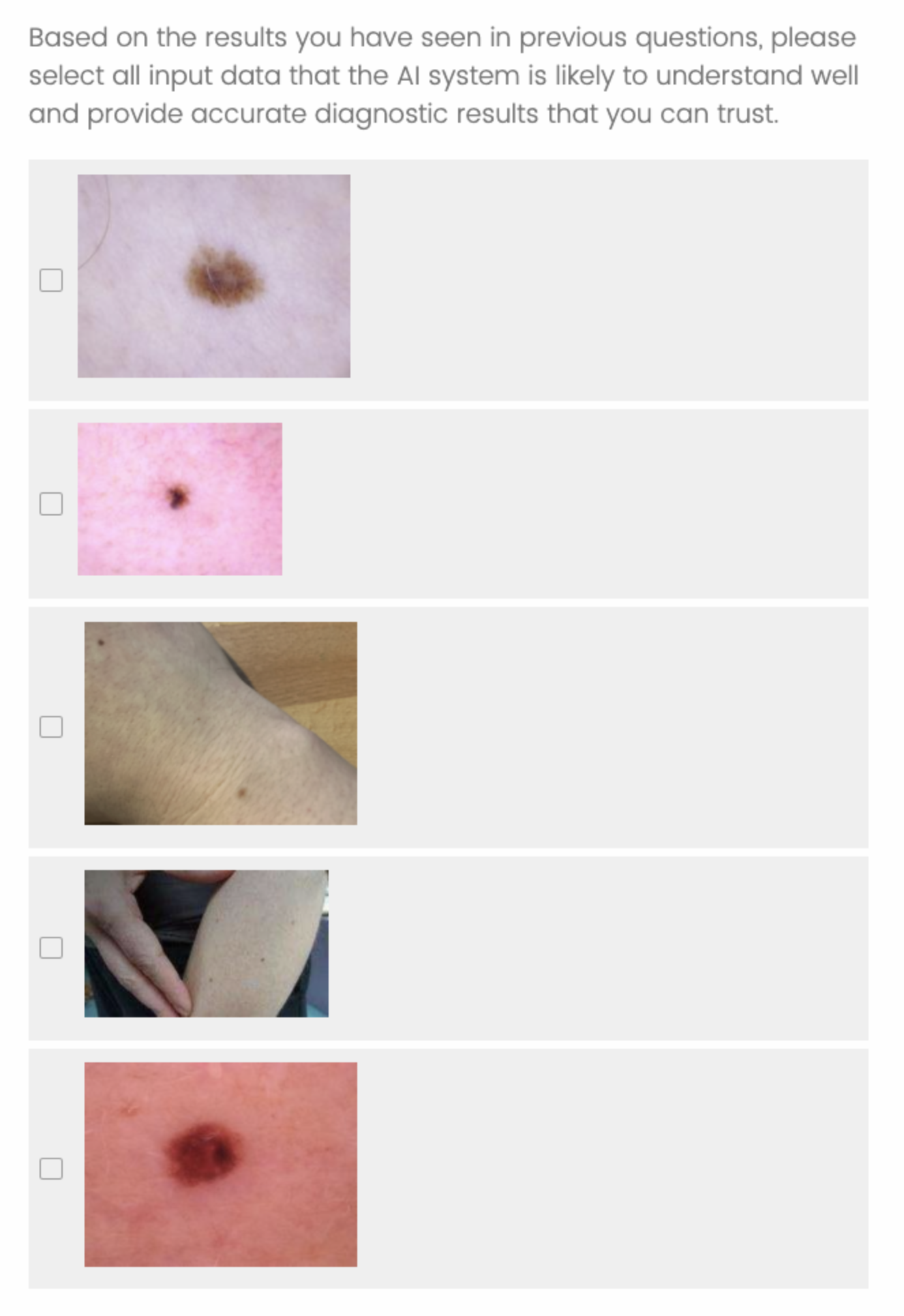}
    \caption{Interface that asks users to select input data that would have high \textsc{confidence score} to explore potential learning effect through their interaction with \textsc{confidence score}.} \label{fig:interface-d}
    \end{subfigure}\hfill
    \caption{List of user study interface. This shows an example interface for skin cancer classifier.} \label{fig:consent}
\end{figure}

\newpage

\begin{figure}[h!]
    \centering
    \begin{subfigure}{0.600\linewidth}
    \includegraphics[width=\linewidth]{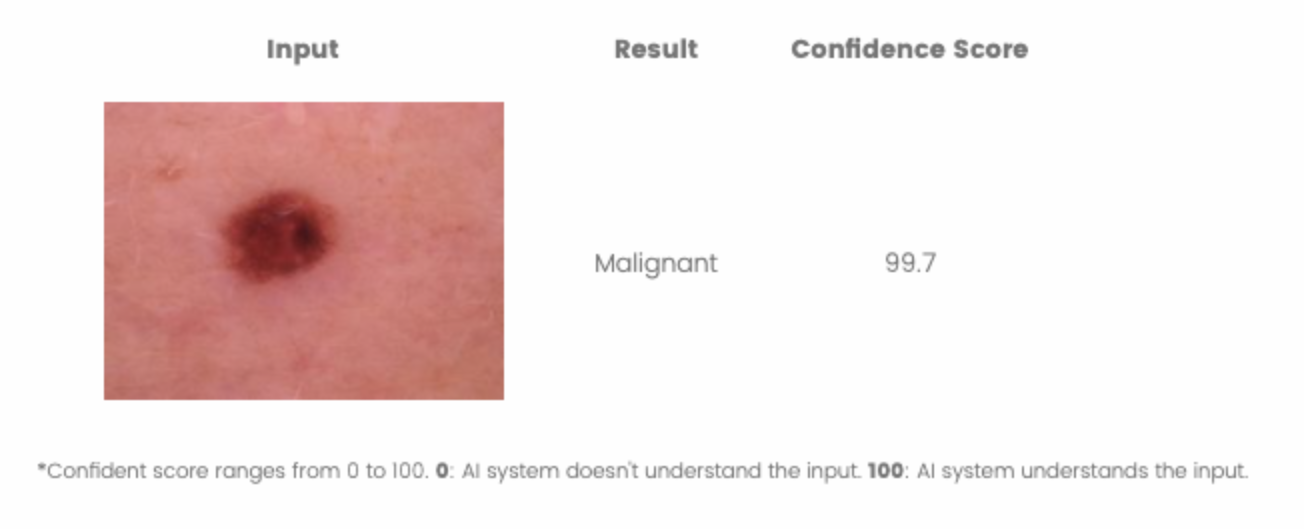}
    \caption{Image input is shown in a visible size.} \label{fig:interface-a}
    \end{subfigure}\hfill

    \medskip
    
    \begin{subfigure}{0.600\linewidth}
    \includegraphics[width=\linewidth]{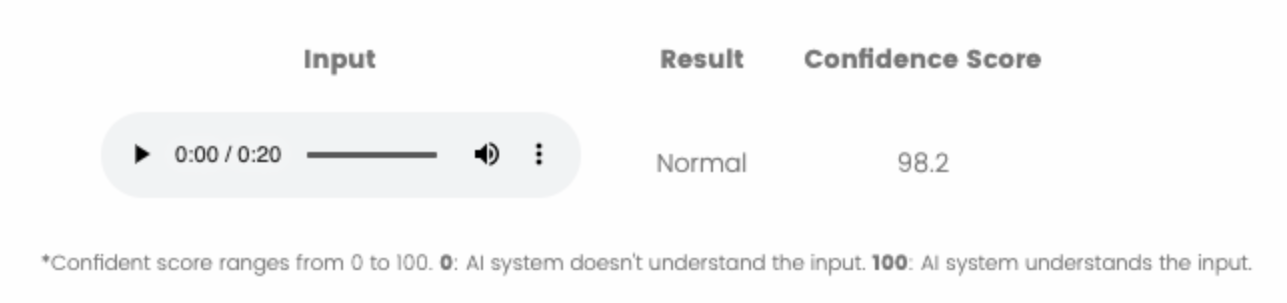}
    \caption{Audio player is embedded for the participants to listen to the input data.} \label{fig:interface-b}
    \end{subfigure}\hfill
    
    \medskip
    
    \begin{subfigure}{0.600\linewidth}
    \includegraphics[width=\linewidth]{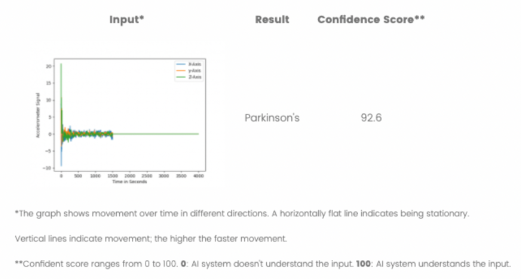}
    \caption{Motion data is shown as a time-series plot of accelerometer signal. We provide additional explanation about how to interpret the signal.} \label{fig:interface-c}
    \end{subfigure}
    \caption{Interface to display different input data types.} \label{fig:interface-modality}
\end{figure}

\begin{figure}[h!]
    \centering
    \begin{subfigure}{0.300\linewidth}
    \begin{subfigure}{0.500\linewidth}
    \includegraphics[width=\linewidth]{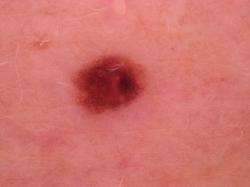}
    \end{subfigure}\hfill
    \begin{subfigure}{0.500\linewidth}
    \includegraphics[width=\linewidth]{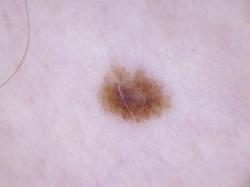}
    \end{subfigure}
    \medskip
    \begin{subfigure}{0.500\linewidth}
    \includegraphics[width=\linewidth]{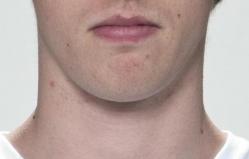}
    \end{subfigure}\hfill
    \begin{subfigure}{0.500\linewidth}
    \includegraphics[width=\linewidth]{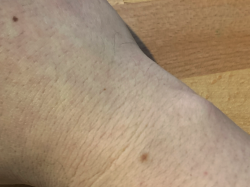}
    \end{subfigure}
    \caption{Input examples for skin cancer classifier for the participants who self-report to have light-colored skin tone.} \label{fig:interface-a}
    \end{subfigure}\hfill
    \begin{subfigure}{0.300\linewidth}
    \begin{subfigure}{0.500\linewidth}
    \includegraphics[width=\linewidth]{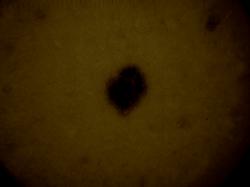}
    \end{subfigure}\hfill
    \begin{subfigure}{0.500\linewidth}
    \includegraphics[width=\linewidth]{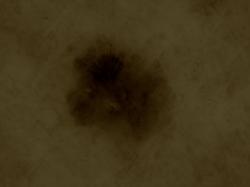}
    \end{subfigure}
    \medskip
    \begin{subfigure}{0.500\linewidth}
    \includegraphics[width=\linewidth]{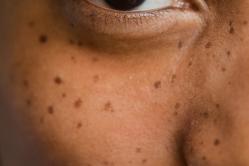}
    \end{subfigure}\hfill
    \begin{subfigure}{0.500\linewidth}
    \includegraphics[width=\linewidth]{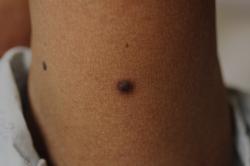}
    \end{subfigure}
    \caption{Input examples for skin cancer classifier for the participants who self-report to have dark-colored skin tone.} \label{fig:interface-a}
    \end{subfigure}\hfill
    \begin{subfigure}{0.300\linewidth}
    \begin{subfigure}{0.500\linewidth}
    \includegraphics[width=\linewidth]{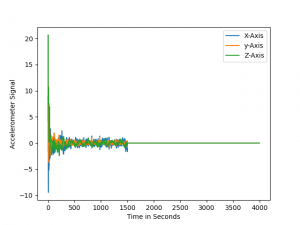}
    \end{subfigure}\hfill
    \begin{subfigure}{0.500\linewidth}
    \includegraphics[width=\linewidth]{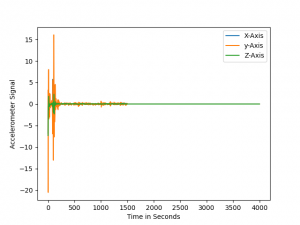}
    \end{subfigure}
    \medskip
    \begin{subfigure}{0.500\linewidth}
    \includegraphics[width=\linewidth]{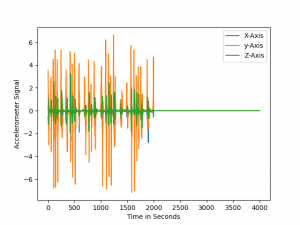}
    \end{subfigure}\hfill
    \begin{subfigure}{0.500\linewidth}
    \includegraphics[width=\linewidth]{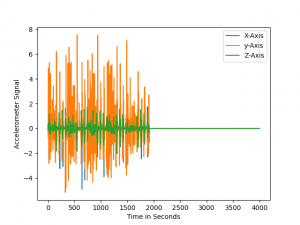}
    \end{subfigure}
    \caption{Input examples for Parkinson's disease classifier.} \label{fig:interface-a}
    \end{subfigure}\hfill
    \caption{List of input examples used in the user study. For each input type, top row shows in-distribution inputs and bottom row shows out-of-distribution inputs. Left column shows positive diagnostic results and right column shows negative diagnostic results. Note that audio samples are not included due to its difficulty to visualize.} \label{fig:interface-modality}
\end{figure}

\section{Performance Metrics}
\label{sec:perf-metrics}
%\cj{detailed definition and equations of TNR @ TPR 95, AUROC, detection accuracy}
In out-of-distribution performance evaluation in Section~\ref{sec:ood-study}, we use the following metrics that has been used in previous out-of-distribution work~\cite{lee2018, sastry2020detecting}:
\begin{itemize}
    \item \textbf{True negative rate (TNR) at 95\% true positive rate} (TPR) is defined as the percentage of correctly detected out-of-distribution samples, when 95\% of in-distribution samples are correctly detected. TNR is calculated $TNR = TN/(FP+TN)$ and $TPR = TP/(TP+TN)$, where TP, TN, FP, and FN are true positive, true negative, false positive, and false negative, respectively.
    \item \textbf{Area under the receiver operating curve (AUROC)} is defined as the area under the plot of true positive rate (TPR) versue false positive rate (FPR), where $TPR = TP/(TP+FN)$ and $FPR = FP/(FP+TN)$. 
    \item \textbf{Detection accuracy} is defined as the maximum classification accuracy over all possible thresholds in classifying in- and out-of-distribution data.
\end{itemize}

\section{Energy-Based OOD Detection Analysis}
\label{sec:energy-analysis}
\begin{figure}[h!]
    \centering
    \begin{subfigure}{0.320\linewidth}
    \includegraphics[width=\linewidth]{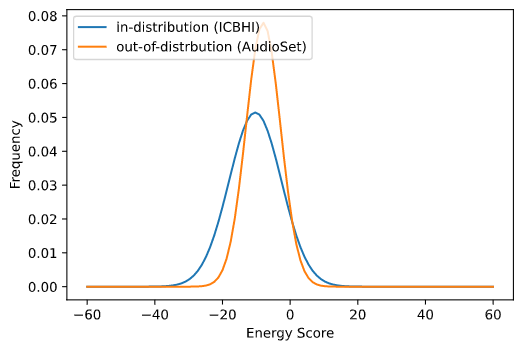}
    \caption{TNR@TPR95: 8.12}
    \label{fig:my_label}
    \end{subfigure}
    \begin{subfigure}{0.320\linewidth}
    \includegraphics[width=\linewidth]{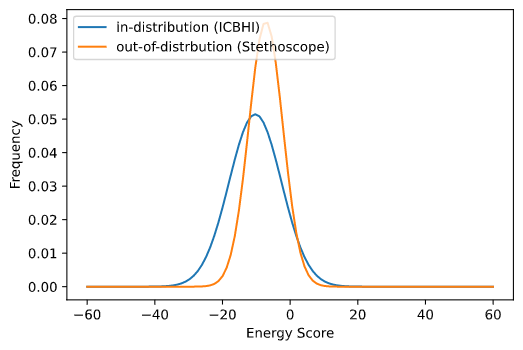}
    \caption{TNR@TPR95: 7.29}
    \label{fig:my_label}
    \end{subfigure}
    
    \begin{subfigure}{0.320\linewidth}
    \includegraphics[width=\linewidth]{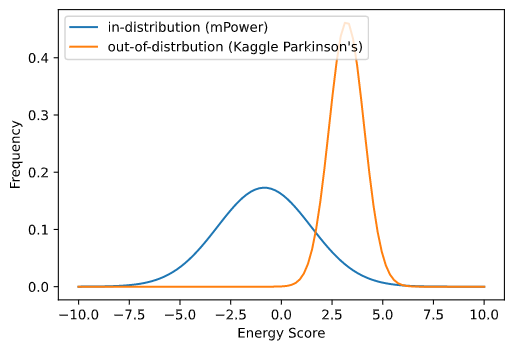}
    \caption{TNR@TPR95: 70.00}
    \label{fig:my_label}
    \end{subfigure}
    \begin{subfigure}{0.320\linewidth}
    \includegraphics[width=\linewidth]{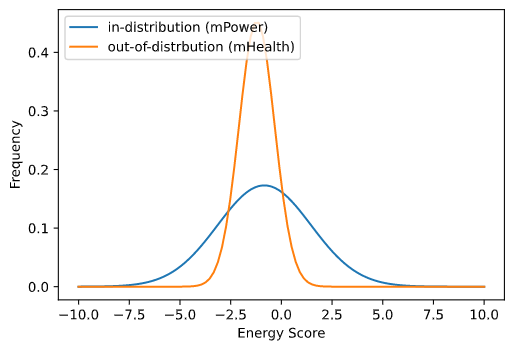}
    \caption{TNR@TPR95: 0.00}
    \label{fig:my_label}
    \end{subfigure}
    \begin{subfigure}{0.320\linewidth}
    \includegraphics[width=\linewidth]{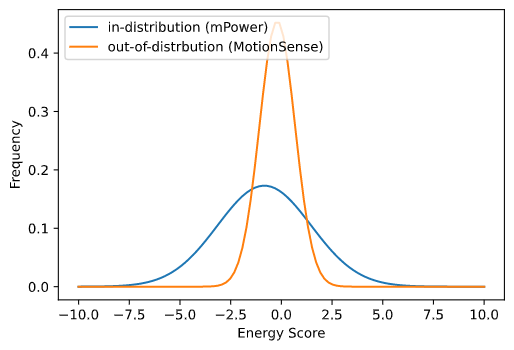}
    \caption{TNR@TPR95: 0.00}
    \label{fig:my_label}
    \end{subfigure}
    
    \begin{subfigure}{0.320\linewidth}
    \includegraphics[width=\linewidth]{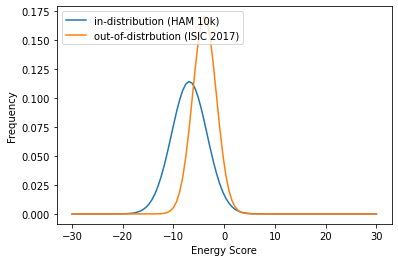}
    \caption{TNR@TPR95: 14.28}
    \label{fig:my_label}
    \end{subfigure}
    \begin{subfigure}{0.320\linewidth}
    \includegraphics[width=\linewidth]{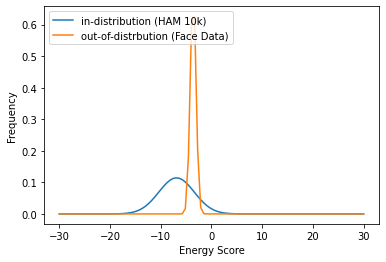}
    \caption{TNR@TPR95: 0.00}
    \label{fig:my_label}
    \end{subfigure}
    \begin{subfigure}{0.320\linewidth}
    \includegraphics[width=\linewidth]{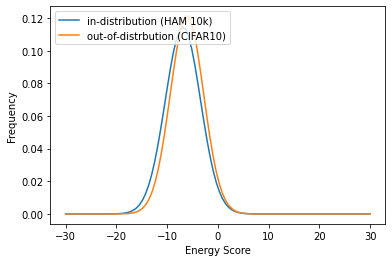}
    \caption{TNR@TPR95: 5.06}
    \label{fig:my_label}
    \end{subfigure}
    
    \caption{Energy score distribution across different in- and out-of-distribution datasets.}
    \label{fig:energy-distribution}
\end{figure}

In Section~\ref{sec:ood-study}, energy-based out-of-distribution detection method does not show comparable performance to methods using Mahalanobis distance and Gram matrices. We further analyze the method by comparing the distribution of energy score between in- and out-of-distribution as shown Figure~\ref{fig:energy-distribution}. In most cases, the distribution of the energy scores are overlapped, making it difficulty to detect out-of-distribution samples using energy score. \revise{In this work, we use energy-based method without fine-tuning, which is suitable for adopting the method to any pre-trained models. However, as the authors have demonstrated in their paper~\cite{energy2020}, fine-tuned energy-based method that requires re-training of a classifier, shows significant improvement in detecting out-of-distribution samples.}

\newpage
\subsection{Out-of-Distribution Performance with Confidence Interval}
\label{sec:model-performance-ci}
\begin{table}[h!]\centering
\resizebox{\textwidth}{!}{
\begin{tabular}{lrrrrrr}\toprule
\textbf{} &\textbf{} &\textbf{} & \textbf{} &\multicolumn{3}{c}{\textbf{Validation on OOD Samples (TNR @ TPR95/AUROC/Detection Accuracy)}} \\\cmidrule{5-7}
\textbf{Health ML Models} &\textbf{In-Distribution} &\textbf{Out-of-Distribution} & \textbf{Distribution Shift} &\textbf{Mahalanobis} &\textbf{Gram} &\textbf{Energy} \\\midrule
%Skin Cancer &HAM10000  &HAM10000 Test Set &4.57 / 48.69 / 50.43 &0.000 / 18.063 / 55.183 &8.069 / 57.565 / 56.369 \\
Skin Lesion &HAM10000  &ISIC 2017 & Covariate/label shift & 10.13 / 58.21 / 59.28 &25.90 / 81.14 / 74.98 & 14.28 / 76.20 / 70.76 \\
(DenseNet-121) & & & &$\pm$2.61/ $\pm$3.30 / $\pm$2.38 & $\pm$1.22 / $\pm$1.89 / $\pm$1.12 & $\pm$0.49 / $\pm$0.18 / $\pm$0.16 \\
% & &ISIC 2016 &4.67 / 51.49 / 53.66 &10.06 / 58.93 / 58.30 & 7.80 / 54.60 / 53.97 \\
%& & & $\pm$0.56 / $\pm$2.66 / $\pm$1.99 & $\pm$0.30 / $\pm$0.37 / $\pm$0.40 &$\pm$0.50 / $\pm$0.64 / $\pm$0.53 \\
& &Face& Open-set recognition  & 100.00 / 99.98 / 99.96 &95.01 / 98.20 / 96.34 & 0.00 / 80.45 / 84.81 \\
& & & & $\pm$0.00 / $\pm$0.02 / $\pm$0.04 & $\pm$1.48 / $\pm$0.41 / $\pm$0.63 & $\pm$0.00 / $\pm$0.14 / $\pm$0.25 \\
& &CIFAR10& Open-set recognition &99.83 / 99.90 / 99.61 &95.14 / 98.66 / 96.90 & 5.06 / 58.33 / 57.89 \\
& & & & $\pm$0.18 / $\pm$0.10 / $\pm$0.39 & $\pm$1.43 / $\pm$1.37 / $\pm$1.94 &$\pm$0.26 / $\pm$0.92 / $\pm$0.67 \\\midrule
%Lung Sound &ICBHI &ICBHI Test Set &4.88 / 49.02 / 50.43 &5.035 / 49.578 / 50.236 &5.194 / 53.454 / 54.238 \\
Lung Sound &ICBHI &AudioSet& Open-set recognition & 97.96 / 99.47 / 97.34 & 96.55 / 99.18 / 95.97 &8.12 / 56.79 / 57.13 \\
(ResNet-34) & & & & $\pm$0.73 / $\pm$0.26 / $\pm$0.45 & $\pm$1.67 / $\pm$0.30 / $\pm$0.62 & $\pm$0.24 / $\pm$0.15 / $\pm$0.14 \\
 & & Stethoscope& Covariate/label shift & 45.60 / 86.27 / 80.57 & 41.77 / 83.75 / 76.05 &  7.29 / 60.98 / 58.94  \\
& & & & $\pm$4.95 / $\pm$1.42 / $\pm$1.55 & $\pm$1.62 / $\pm$0.63 / $\pm$0.38 & $\pm$1.22 / $\pm$0.74 / $\pm$0.63 \\\midrule
%Parkinson's Disease &mPower &mPower Test Set &3.75 / 50.47 / 51.33 &19.418 / 36.309 / 61.995 &4.800 / 50.794 / 51.016 \\
Parkinson's Disease &mPower &MotionSense& Open-set recognition &100.00 / 99.86 / 99.89 &100.00 / 99.94 / 99.60 & 0.00 / 58.71 / 64.96 \\
(5$\times$1D-Conv) & & & & $\pm$0.00 / $\pm$ 0.13 / $\pm$0.10 & $\pm$0.00 / $\pm$0.24 / $\pm$0.14 & $\pm$0.00 / $\pm$0.59 / $\pm$0.32 \\
 & &mHealth& Open-set recognition &100.00 / 100.00 / 100.00 &100.0 / 99.99 / 99.99 & 0.00 / 41.41 / 59.44 \\
& & & &$\pm$ 0.00 / $\pm$0.00 / $\pm$0.00 & $\pm$0.00 / $\pm$0.02 / $\pm$0.01 &$\pm$0.00 / $\pm$1.09 / $\pm$1.10 \\
& &Kaggle Parkinson's& Covariate/label shift& 98.00 / 99.89 / 99.47 & 98.96 / 99.96 / 99.67 & 70.00 / 95.91 / 93.34 \\
& & & & $\pm$2.45 / $\pm$0.14 / $\pm$1.25 & $\pm$0.00 / $\pm$0.02 / $\pm$0.03 &$\pm$4.68 / $\pm$0.30 / $\pm$0.32 \\\bottomrule
\end{tabular}}
\caption{Out-of-Distribution Detection Performance Across Multiple Tasks. Evaluation is repeated for 5 times. Mean and standard deviation of metrics are reported. }
\label{tab:ood-performance-ci}
\end{table}

\end{document}